\documentclass[journal ]{new-aiaa}
\usepackage[utf8]{inputenc}
\usepackage{textcomp}

\usepackage[dvipsnames]{xcolor}
\definecolor{lightblue}{rgb}{0.63, 0.74, 0.78}
\definecolor{seagreen}{rgb}{0.18, 0.42, 0.41}
\definecolor{orange}{rgb}{0.85, 0.55, 0.13}
\definecolor{silver}{rgb}{0.69, 0.67, 0.66}
\definecolor{rust}{rgb}{0.72, 0.26, 0.06}
\definecolor{purp}{RGB}{68, 14, 156}

\colorlet{lightrust}{rust!50!white}
\colorlet{lightorange}{orange!25!white}
\colorlet{lightlightblue}{lightblue}
\colorlet{lightsilver}{silver!30!white}
\colorlet{darkorange}{orange!75!black}
\colorlet{darksilver}{silver!65!black}
\colorlet{darklightblue}{lightblue!65!black}
\colorlet{darkrust}{rust!85!black}
\colorlet{darkseagreen}{seagreen!85!black}

\usepackage{framed}
\usepackage{physics}
\usepackage{graphicx}
\usepackage{amsmath}
\usepackage[version=4]{mhchem}
\usepackage{siunitx}
\usepackage{longtable,tabularx}
\usepackage{bm}
\usepackage[ruled, vlined]{algorithm2e}
\usepackage{booktabs}
\usepackage{multirow}
\usepackage{tabularx}
\usepackage{ragged2e}
\usepackage{booktabs, tabularx, enumitem}
\newcolumntype{Y}{>{\raggedright\arraybackslash}X}

\usepackage{color,soul}
\usepackage{microtype}
\usepackage{tikz}
\usepackage{lipsum} 
\usepackage{graphicx}
\usepackage{float}
\usepackage[nameinlink]{cleveref}

\usepackage{amssymb}
\usepackage{physics}
\usepackage{xcolor}

\newcommand{\bfsym}[1]{\boldsymbol{#1}}

\usepackage{hyperref}
\hypersetup{
    colorlinks = true,
    linkcolor  = blue,
    urlcolor   = blue,
    citecolor  = blue,
}

\usepackage{epstopdf}
\AppendGraphicsExtensions{.tif}

\title{Data‑Driven Reduced‑Complexity Modeling of Fluid Flows: \\ A Community Challenge}

\author{\hspace{7mm}Oliver T. Schmidt\footnote{Co-first authors.
}\footnote{Associate Professor, Department of Mechanical and Aerospace Engineering, Senior Member AIAA.}
\hspace{1.9cm} \& \hspace{1.9cm}
Aaron Towne$^{\text{*}}$\footnote{Associate Professor, Department of Mechanical Engineering, Senior Member AIAA.}
}
\affil{University of California San Diego, La Jolla, CA
\hspace{1cm}
University of Michigan, Ann Arbor, MI}

\author{Adrian Lozano-Duran\footnote{Associate Professor, Graduate Aerospace Laboratories, Senior Member AIAA.}}
\affil{California Institute of Technology, Pasadena, CA}

\author{Scott T. M. Dawson\footnote{Associate Professor, Department of Mechanical, Materials and Aerospace Engineering, AIAA Associate Fellow.}}
\affil{Illinois Institute of Technology, Chicago, IL}

\author{Ricardo Vinuesa\footnote{Associate Professor, Department of Aerospace Engineering.}}
\affil{University of Michigan, Ann Arbor, MI}

\begin{document}

\maketitle

\begin{abstract}
We introduce a community challenge designed to facilitate direct comparisons between data-driven methods for compression, forecasting, and sensing of complex aerospace flows. The challenge is organized into three tracks that target these complementary capabilities: compression (compact representations for large datasets), forecasting (predicting future flow states from a finite history), and sensing (inferring unmeasured flow states from limited measurements). Across these tracks, multiple challenges span diverse flow datasets and use cases, each emphasizing different model requirements. The challenge is open to anyone, and we invite broad participation to build a comprehensive and balanced picture of what works and where current methods fall short. To support fair comparisons, we provide standardized success metrics, evaluation tools, and baseline implementations, with one classical and one machine-learning baseline per challenge. Final assessments use blind tests on withheld data. We explicitly encourage negative results and careful analyses of limitations. Outcomes will be disseminated through an AIAA Journal Virtual Collection and invited presentations at AIAA conferences.
\end{abstract}


\newpage
\tableofcontents

\clearpage

\section{Introduction}
\vspace{\baselineskip}

Driving community progress through well-designed challenges is an established practice in machine learning and has led to major breakthroughs, for example, in image recognition through the ImageNet challenge \cite{deng2009imagenet}, in natural language understanding through the GLUE benchmark and analysis platform \cite{wang2018glue}, and in protein folding through the CASP series of challenges \cite{MOULT2005285}. In aerospace science and fluid dynamics, such efforts are comparibly scarce, but notable examples include the ML4CFD competition on machine learning for airfoil design \cite{yagoubi2024neurips}, the Collaborative Testing Challenge on data-driven turbulence modeling \cite{rumsey2022data}, “The Well” challenge, which assembles a broad collection of physics simulations, many of them fluid-dynamics problems, for machine learning \cite{ohana2024well}, and, very recently, a common task framework for machine learning of dynamical systems \cite{kutz2025acceleratingscientificdiscoverycommon}.

\begin{figure}[H]
    \centering
    \noindent\includegraphics[width=1\textwidth,trim={0 0.75in 0 0},clip]{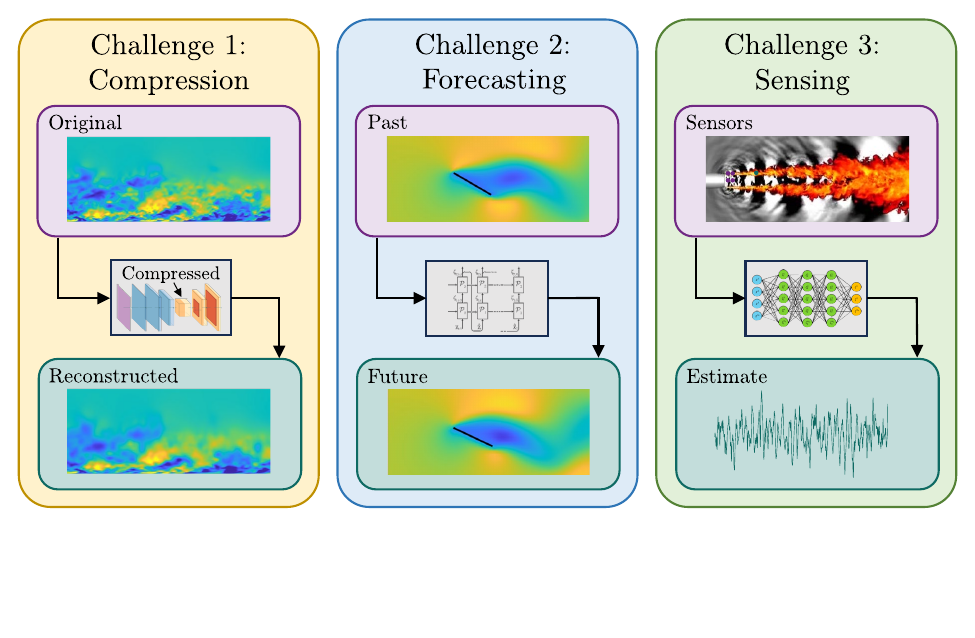}
    \caption{Overview of the challenges, which span the core capabilities of compression, forecasting, and sensing.  Each challenge type conceptually includes input data (plum boxes) and target output data (teal boxes) and seeks methods to perform the desired task (gray boxes). }
    \label{fig:overview}
\end{figure}

This community challenge targets complex aerospace flows across multiple application-driven cases and organizes them into three complementary capabilities: \emph{compression}, \emph{forecasting}, and \emph{sensing}. It pairs standardized evaluation with baseline methods drawn from established statistical tools, dynamical systems, and machine learning.
\emph{Compression} provides compact representations that are essential not only for storage but, crucially, to ensure that other methods remain tractable for large datasets and deployable in near real time. \emph{Forecasting} targets prediction in time, i.e., inferring future flow states from a finite history. \emph{Sensing} targets prediction in space, i.e., inferring unmeasured states or derived quantities from limited measurements. Together, these tasks cover common operational needs in analysis, design, and control.  Each challenge type is represented schematically in Figure~\ref{fig:overview}.

The challenge emphasizes purely data-driven methods. Our goal is to enable fair comparisons using standardized success metrics and common baselines on relevant example data, spanning classical statistical or dynamical approaches and modern machine-learning methods, in a fully open and transparent setting. While each challenge specifies datasets and metrics, participants are completely free in their choice of method so long as it is purely data-driven. 

To anchor comparisons, we provide established baseline methods, one classical and one ML method per challenge, paired with diverse flow datasets, largely drawn from a public database specifically constructed for evaluating reduced-complexity models for fluid flows \citep{towne2023database}. Each dataset is paired with specific challenges that reflect representative aerospace use cases. An overview of the challenges, datasets, and baseline methods is given in Table~\ref{tab:challenge-summary}. 

Participants are expected to release their code and undergo blinded evaluation on held-out data by submitting results to the designated point of contact for each challenge. Dissemination will include a Virtual Issue of the \emph{AIAA Journal} and invited talks in special sessions at AIAA conferences. Participation guidelines are detailed in Section~\ref{sec:participation}, and additional information and periodic updates are available on the challenge webpage: \url{fluids-challenge.engin.umich.edu}.

Our objective is to identify competitive methods, distill best practices, and advance community understanding. Novel architectures are welcome, as are careful adaptations or lightweight customizations of established methods. Importantly, submissions that illuminate limitations or failure modes are also valuable contributions to the community.

\vspace{0.5cm}

\begin{table}[h!]
\small
\centering
\caption{Summary of the three challenges, their sub-challenges, datasets, and baseline methods.}
\label{tab:challenge-summary}
\renewcommand{\arraystretch}{1.12}
\begin{tabularx}{\textwidth}{@{} l p{0.3\textwidth} p{0.22\textwidth} Y @{}}
\toprule
\textbf{Challenge} & \textbf{Sub-challenges} & \textbf{Data} & \textbf{Baseline methods} \\
\midrule
\midrule
\multirow{2}{*}{\textbf{1. Compression}} 
& 1.1 Compression of 2D turbulent boundary layer velocity fields 
& DNS of high-Re turbulent boundary layer   
& Proper orthogonal decomposition (POD) \\
\cmidrule(l){2-3}
& 1.2 Compression of 3D turbulent boundary layer velocity fields 
& DNS of high-Re turbulent boundary layer 
& Convolutional autoencoders (CNN-AE) \\
\midrule
\midrule
\multirow{2}{*}{\textbf{2. Forecasting}} 
& 2.1  Forecasting of a turbulent cavity flow 
& Time-resolved PIV of a cavity 
& Dynamic mode decomposition (DMD) \\
\cmidrule(l){2-3}
& 2.2 Forecasting of pitching airfoils with parametric variations 
& DNS of pitching airfoil across pitching angles and frequencies 
& Long short-term memory network (LSTM) \\
\midrule
\midrule

\multirow{2}{*}{\textbf{3. Sensing}} 
& 3.1 Sensing of a turbulent jet using sparse measurements
& LES of axisymmetric jet 
& Wiener filter (WF) \& Multi-layer perceptron (MLP) \\
\cmidrule(l){2-3}
& 3.2 Sensing of a turbulent boundary layer from wall stress
& DNS of low-Re turbulent boundary layer
& Linear stochastic estimation (LSE) \& convolutional neural network (CNN) \\
\bottomrule
\end{tabularx}
\end{table}


\clearpage

\section{Challenge 1: Compression}\label{sec:compression}

\subsection{Introduction}

Compressing turbulent flow data poses a fundamental challenge due to
the chaotic nature of turbulence, its fractal structure, and the wide
range of spatial and temporal scales involved. These characteristics
make it difficult to reduce the dimensionality of the data without
sacrificing critical physical information. Effective compression must
therefore strike a balance between a compact representation and the
ability to accurately reconstruct key flow features. In practical
scenarios, such as real-time monitoring or on-the-fly flow
compression, the wall time and computational cost associated with
compression and reconstruction are equally important. These
considerations motivate the need for novel modeling strategies that
optimize not only the compression ratio and reconstruction fidelity,
but also the efficiency of the process in terms of cost and time.

We pose two representative challenges within the context of a
turbulent boundary layer: (i) The first focuses on compressing and
reconstructing all three components of the velocity fluctuations on
two-dimensional planes extracted from the flow; (ii) The second
extends this task to the three-dimensional reconstruction of
streamwise velocity fluctuations. These scenarios reflect different
levels of complexity and spatial information. To evaluate performance,
we consider a combination of metrics that capture trade-offs between
reconstruction accuracy, measured via normalized mean square error (NMSE),
the achieved compression ratio, and the computational cost of both
compression and reconstruction.

\subsection{Baseline methods}\label{compression:methods}

Two baseline methods are considered to illustrate the compression challenge: space-only POD with $N$ modes and a convolutional autoencoder (CNN-AE) with a latent space of dimension $N$.

\subsubsection{Proper orthogonal decomposition (POD)}\label{sec:POD}

Proper orthogonal decomposition (POD) constitutes the optimal linear method for data compression \citep{lumley1970stochastic}.  Here, we use the space-only variant of POD \cite{sirovich1987turbulence, towne2018spectral} in which the spatially discretized flow state $\bfsym{q}(t)$ is approximated by $\tilde{\bfsym{q}}(t)$ as a linear combination of $N$ orthogonal modes $\boldsymbol{\phi}_i$ and their corresponding time-dependent coefficients $a_i(t)$, i.e.,
\begin{equation}
\tilde{\bfsym{q}}(t) = \sum_{i=1}^N a_i(t) \, \boldsymbol{\phi}_i.
\label{eq:POD_expansion}
\end{equation}
The optimal spatial modes are obtained from the eigendecomposition of the spatial correlation matrix $\bfsym{C}_{qq} = E\left\{\bfsym{q}\bfsym{q}^{*}\right\}$. The temporal coefficients are recovered by projecting the state $\bfsym{q}$ onto each mode and are used to reconstruct the state according to (\ref{eq:POD_expansion}).  POD has been extensively employed to compress fluid flows, often as a starting point for additional modeling efforts \citep{berkooz1993proper, rowley2017model, colonius2025modal}.

\subsubsection{Convolutional autoencoder (CNN-AE)}

Autoencoders offer a natural nonlinear extension of POD \citep{rumelhart1986learning, hinton2006reducing} and consist of an encoder and decoder. The encoder is a nonlinear function $E_{\theta}$ that maps the flow state $\bfsym{q}$ to a latent representation $\bfsym{a} \in \mathbb{R}^N$ that captures the essential features of the input in a reduced-dimensional space, i.e., $\bfsym{a} = E_{\theta}(\bfsym{q})$.  The decoder reconstructs the flow field from the latent space using another nonlinear function $D_{\theta}$, i.e., $\tilde{\bfsym{q}} = D_{\theta}(\bfsym{a})$, where $\tilde{\bfsym{q}}$ is the reconstructed approximation of the original flow state $\bfsym{q}$. During training, the network parameters $\theta$ are optimized to minimize a desired loss function.

Here, we use a convolutional autoencoder (CNN-AE) \citep{masci2011stacked} with a latent space of dimension $N$ to compress and reconstruct the flow-field data. Both the encoder and decoder are built using convolutional neural networks, allowing the model to leverage the spatial correlations inherent in the flow fields for more accurate and efficient representation and reconstruction.  The network is trained by minimizing the mean-squared reconstruction error.  Autoencoders have become popular in recent years for identifying features and low-dimensional subspaces in fluid flows \citep{lee2020model, solera2024beta}.

\subsection{Challenge 1.1: Compression of 2D turbulent boundary layer velocity fields}

\vspace{-0.25cm}
\begin{figure}[H]
 \centering
 \includegraphics[width=1\textwidth,trim={0 1.5in 0 0},clip]{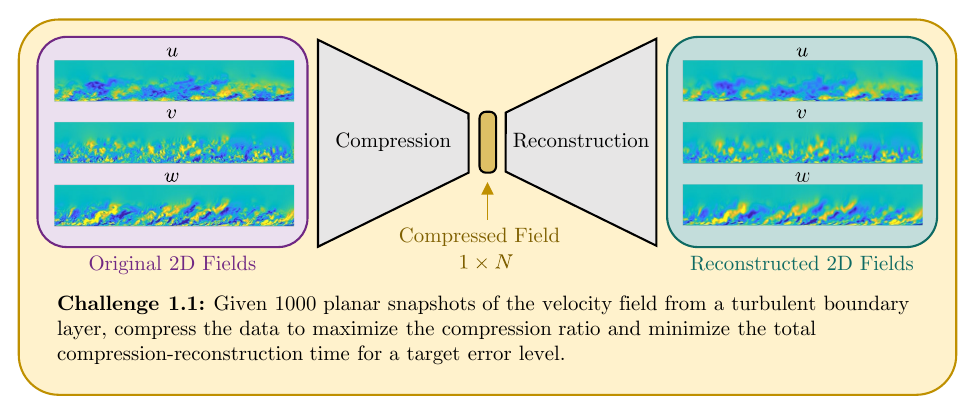}
\end{figure}
\vspace{-0.25cm}
\subsubsection{Data}\label{compression:data}

This challenge uses data from a direct numerical simulation (DNS) of a zero-pressure-gradient incompressible turbulent boundary layer.  In particular, we use data from the BL2 case described in Ref.~\citep{towne2023database}.  The full database contains 7500 three-dimensional flowfields sampled every $\Delta t^{+} \approx 4$ and spanning a physical time of 7.4 eddy turnovers. The training dataset for this challenge includes 5,000 planar snapshots in the streamwise–wall-normal ($x$–$y$) plane, covering friction Reynolds numbers in the range $Re_{\tau} \approx 480\text{--}1020$. Each snapshot contains the streamwise ($u$), wall-normal ($v$), and spanwise ($w$) velocity fluctuations on a $4000 \times 148$ grid. The test set comprises an additional 1,000 snapshots drawn from the same simulation.

\subsubsection{Baseline results}
\label{compression:challenge5_1}

The performance of the methods in the compression challenge is assessed by
examining the compression ratio and compression/reconstruction times,
as functions of the root-normalized-mean-squared error (root-NMSE) 
obtained using compressor--reconstructor
pairs with varying numbers of retained modes / latent-space dimensions
($N$).  The normalized mean-squared error (NMSE) is defined as
\begin{equation}
e = \frac{E\left\{ \|  \tilde{\boldsymbol{q}}(t) - \boldsymbol{q}(t) \|^2 \right\}}{E\left\{ \|  \boldsymbol{q}(t) \|^2 \right\}},
\label{eq:NMSE_compression}
\end{equation}
where the expectation is a time average over snapshots and the norm constitutes spatial integration over the field.

POD is conducted on the three velocity components for 11
cases, using retained mode counts ranging from $N=4$ up to $4096$,
increasing in successive powers of two.  The 2D-CNN-AE comprises three
channels (each with one velocity component) are stacked into a 4D
array. The encoder consists of two convolutional blocks, each with a
3x3 convolution (16 and 32 filters, respectively), ReLU activation,
and 2x2 max-pooling, reducing spatial dimensions by 4. The bottleneck
uses a fully connected layer to produce a latent vector of size
$N$. The decoder uses a fully connected layer to reshape the latent
representation, followed by two transposed convolutional layers (4x4,
stride 2) and a final 3x3 convolution to reconstruct the 3-channel
output, trained with Adam optimization for 100 epochs using a mean-squared-error loss function.

\begin{figure}[hbt!]
 \centering
 \includegraphics[width=1\textwidth]{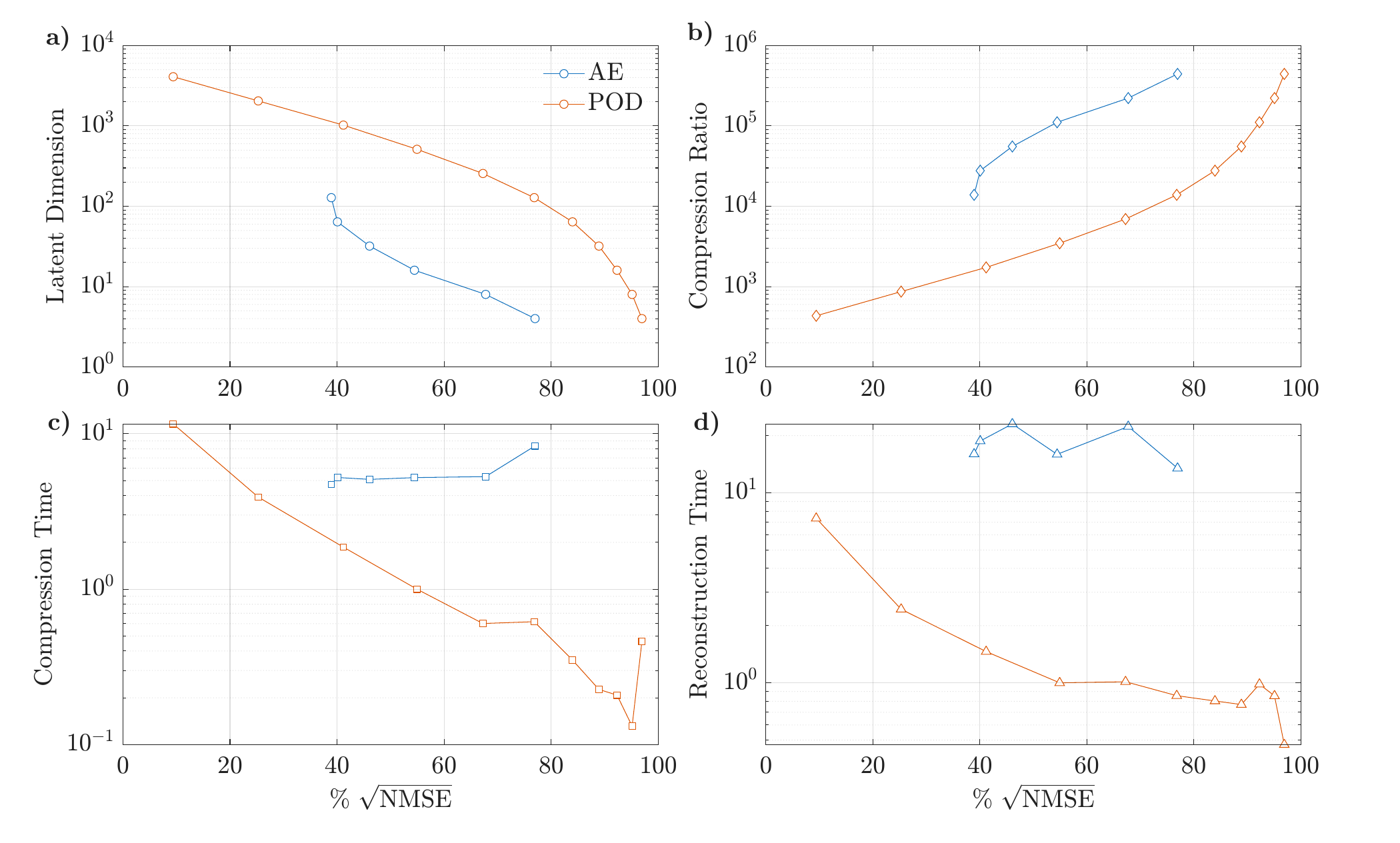}
 \caption{Challenge 1.1: Application of compression metrics of success
   in 2-D planes of the streamwise velocity in a turbulent boundary
   layer using 2D-CNN-AE and POD. The compression and
   reconstruction times are normalized using the POD method with 512
   modes.\label{fig:compression:results_2D}}
 \end{figure}

\begin{figure}[hbt!]
 \centering
 \includegraphics[width=1\textwidth]{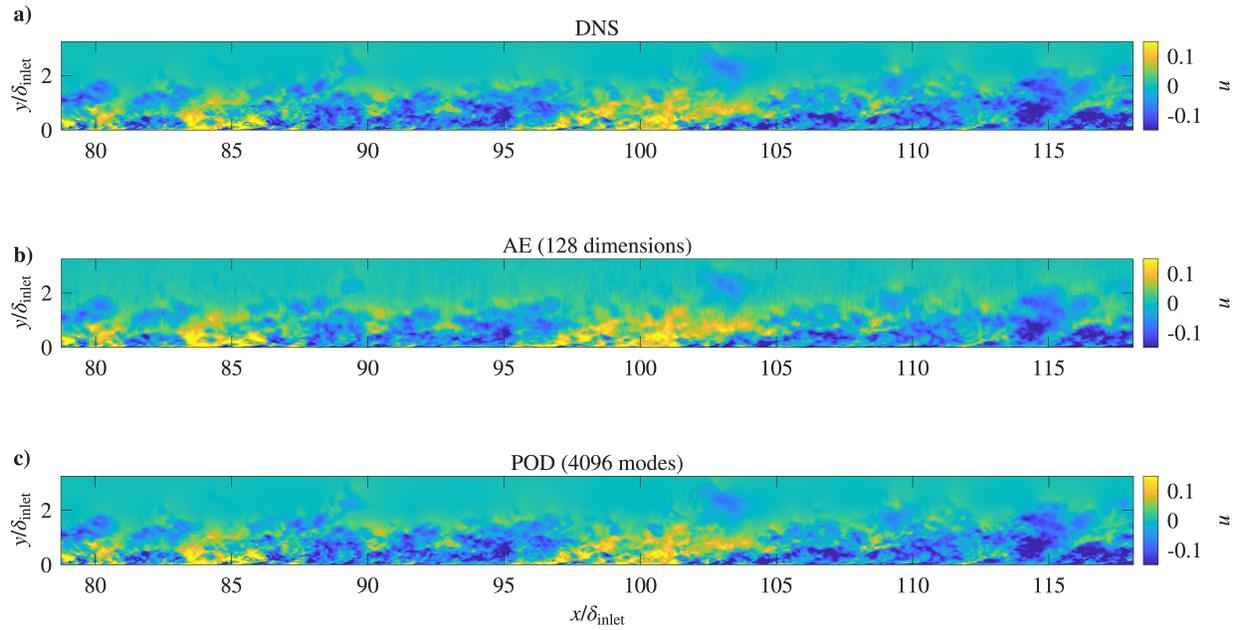}
 \caption{Challenge 1.1: Application of compression to 2-D
   fields. Example snapshot of the streamwise velocity field (a),
   its reconstruction using a 2D-CNN-AE with a latent dimension of 128
   (b), and reconstruction using 4096 POD modes (c). The
   streamwise ($x$) and wall-normal ($y$) coordinates are
   nondimensionalized by the boundary layer thickness at the inlet of
   the simulation
   ($\delta_\text{inlet}$). \label{fig:compression:example_2D} }
 \end{figure}

The example results for Challenge 1.1 are shown in
Figure~\ref{fig:compression:results_2D}, which presents a comparative
analysis of compressing and reconstructing the three components of the
turbulent velocity fluctuations using POD and CNN-AE.  An example of
the reconstructed $u$ is shown in
Figure~\ref{fig:compression:example_2D}.

Figure~\ref{fig:compression:results_2D}(a) shows the relationship between the root-NMSE and the dimension of the latent space (or the number of POD modes). This plot
illustrates the fundamental trade-off between dimensionality reduction
and reconstruction fidelity. Lower root-NMSE values at smaller latent
dimensions reflect a more effective compression method, capable of
capturing essential flow features in a compact representation. While
latent-space dimensionality is not used as a standalone performance
metric (due to method-dependent definitions) we include root-NMSE versus
latent-dimension plots for both POD (interpreted as the number of
modes) and 2D-CNN-AE, as they offer valuable insights.  For both POD
and 2D-CNN-AE, the root-NMSE consistently decreases with increasing latent
dimension, indicating improved reconstruction accuracy at higher
representation capacity. However, to reach a comparable root-NMSE, POD
requires significantly larger latent spaces than 2D-CNN-AE. This
highlights 2D-CNN-AE's superior ability to represent the complex,
nonlinear structures in turbulent flows, in contrast to the linear
basis functions used in POD.

Figure~\ref{fig:compression:results_2D}(b) shows the root-NMSE as a function of the compression ratio. In this analysis, we consider only the size of the compressed
(latent) representation, excluding the overhead associated with the
compression and reconstruction machinery, such as neural network
parameters or modal basis functions. As a result, the reported
compression ratio reflects the asymptotic limit of \emph{infinite
data}, where the size of the compression algorithm itself becomes
negligible compared to the size of the data.  Both POD and 2D-CNN-AE
exhibit a gradual increase in root-NMSE with increasing compression,
reflecting the expected tradeoff between data reduction and
reconstruction accuracy. For a given root-NMSE, 2D-CNN-AE achieves
significantly higher compression ratios than POD, highlighting its
efficiency in representing turbulent flows. This result suggests that
2D-CNN-AE can reduce data volume more aggressively without
compromising reconstruction fidelity---an important advantage for
applications that demand high compression.

Figure~\ref{fig:compression:results_2D}(c) and (d) examine root-NMSE as a function of compression and reconstruction time, respectively.  To reduce variability introduced by hardware differences, both compression and reconstruction times are reported relative to a baseline method executed on the same hardware
used for each submission. In this case, both compression and
reconstruction times are normalized by the corresponding values
obtained using the POD method with 512 modes.  For POD, both times
increase as the root-NMSE decreases, reflecting the additional computational
effort required to achieve higher accuracy due to the larger number of
retained modes. In contrast, the compression and reconstruction times
for 2D-CNN-AE are generally higher than those for POD over the range
of root-NMSE values considered and remain relatively constant, regardless
of the desired accuracy.

In summary, the 2D-CNN-AE outperforms POD in terms of compression ratio, 
achieving higher reconstruction accuracy for smaller compressed data sizes. 
However, the 2D-CNN-AE shows signs of saturation as the latent-space dimension 
increases and never yields errors below 40\%. In contrast, POD achieves errors 
as low as 10\%. POD also offers faster compression and reconstruction, 
outperforming the 2D-CNN-AE in computational efficiency over the root-NMSE range  
considered. Notably, the compression-reconstruction time for 2D-CNN-AE remains 
relatively constant as the target root-NMSE decreases, indicating that 2D-CNN-AE 
may surpass POD in efficiency as well for lower root-NMSE values.

\subsection{Challenge 1.2: Compression of 3D turbulent boundary layer velocity fields}
\label{compression:challenge5_2}
\vspace{-0.25cm}
\begin{figure}[H]
 \centering
 \includegraphics[width=1\textwidth,trim={0 1.5in 0 0},clip]{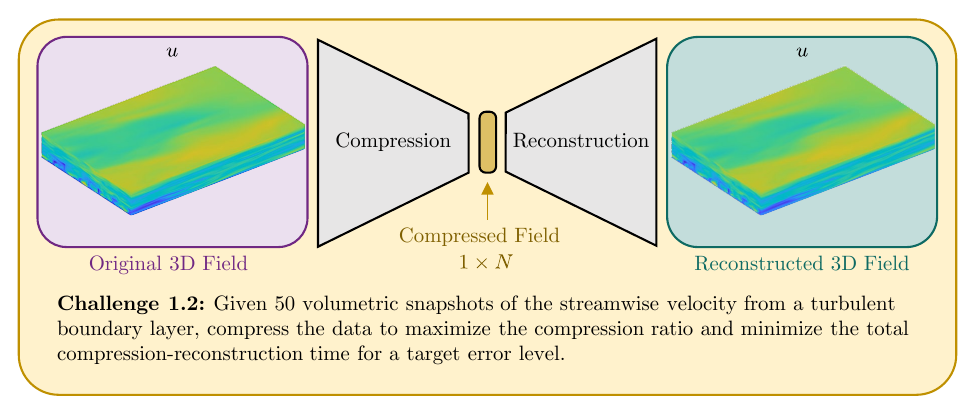}
\end{figure}
\vspace{-0.25cm}

\subsubsection{Data}

This challenge uses data from the same DNS of a zero-pressure-gradient incompressible turbulent boundary layer~\citep{towne2023database}. The training set consists of 500 volumetric (three-dimensional) snapshots of streamwise velocity fluctuations, extracted from a subdomain of the boundary layer with $Re_{\tau} \approx 920\text{--}945$.  Each volumetric field is defined on a $256 \times 64 \times 256$ grid, corresponding to the streamwise, wall-normal, and spanwise directions, respectively. The test set includes 50 additional snapshots of the volumetric field.

\subsubsection{Baseline results}
 
POD is performed on the full 3-D streamwise velocity field for four
cases, with the number of retained POD modes set to $N=50$, $100$,
$200$, and $400$. The 3D-CNN-AE consists of an encoder with five
blocks, each containing a 3-D convolution (3x3x3 filters, increasing
from 16 to 256 channels), batch normalization, ReLU activation, and
2x2x2 average pooling, reducing spatial dimensions to [8, 2, 8]. The
bottleneck layer applies a 3x3x3 convolution to produce a latent
representation with $N$ channels. The decoder mirrors the encoder with
five blocks of transposed convolutions (4x4x4, stride 2) to upsample,
followed by convolution, batch normalization, and ReLU, ending in a
final 3x3x3 convolution to output a single-channel reconstructed
volume.  The 3D-CNN-AE is trained with Adam optimization for 650
epochs using mean-squared-error loss.

\begin{figure}[hbt!]
 \centering
 \includegraphics[width=1\textwidth]{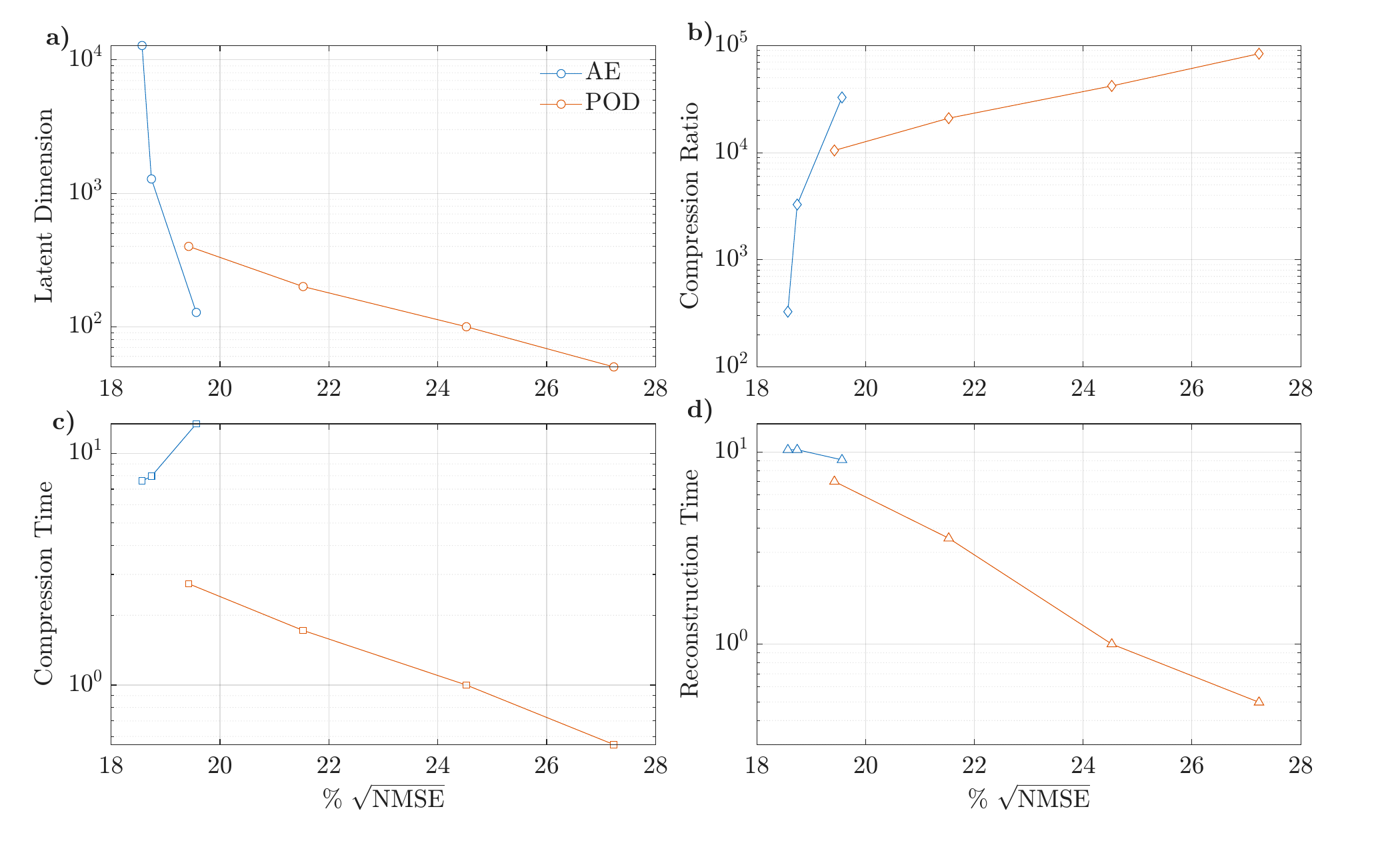}
 \caption{Challenge 1.2: Application of compression: metrics of
   success in 3-D fields of the streamwise velocity fluctuations in a
   turbulent boundary layer using 3D-CNN-AE and
   POD. The compression and reconstruction times are normalized using
   the POD method with 100 modes.\label{fig:compression:results_3D}}
 \end{figure}

\begin{figure}[hbt!]
 \centering
 \includegraphics[width=1\textwidth]{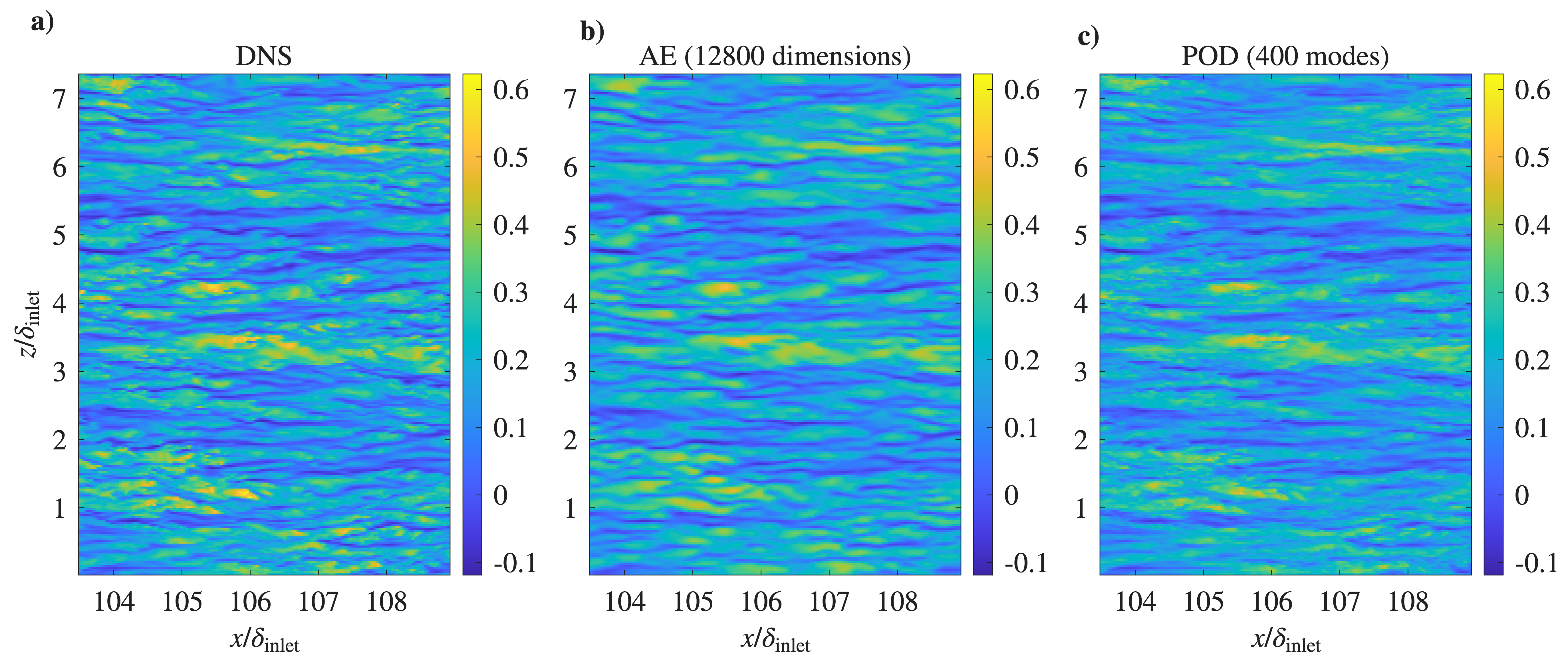}
 \caption{Challenge 1.2: Application of compression to 3-D fields:
   Example of a 2-D cut extracted from the 3-D field of the streamwise
   velocity fluctuations (a), its reconstruction using a 3D-CNN-AE
   with a latent dimension of 12800 (b) and reconstruction using
   400 POD modes (c). The streamwise ($x$) and spanwise ($z$)
   coordinates are nondimensionalized by the boundary layer thickness
   at the inlet of the simulation
   ($\delta_\text{inlet}$). \label{fig:compression:example_3D} }
\end{figure}

The results for Challenge 1.2 are shown in
Figure~\ref{fig:compression:results_3D}, which presents a comparative
analysis of compressing and reconstructing the 3-D field for the
streamwise velocity fluctuations using POD and 3D-CNN-AE.  An example
of the 3-D reconstructed $u$ is shown in
Figure~\ref{fig:compression:example_3D}. The results for the 3-D case exhibit trends that differ from those in the 2-D case,  and the superiority of the 3D-CNN-AE is less pronounced. Although the 3D-CNN-AE yields slightly lower reconstruction 
errors, its overall behavior more closely mirrors that of POD. For example, 
both 3D-CNN-AE and POD achieve reconstruction errors of about $20\%$ at 
compression ratios on the order of $10^4$.

A clear limitation of POD is its restriction to a maximum of 500 modes
due to the method of snapshots. This constraint arises because the
number of available POD modes is bounded by the number of training
flow fields. In contrast, the 3D-CNN-AE does not suffer from this
limitation, as its latent dimension can be increased independently of
the number of training samples.

In all cases, the compression-reconstruction time remains on the order
of a few seconds for both the 3D-CNN-AE and POD. Interestingly, the
reconstruction error of the 3D-CNN-AE does not improve substantially
when increasing the latent dimension from 128 to 12,800. This suggests
that the performance bottleneck of the current architecture lies not
in the latent dimensionality, but rather in the convolutional layers.

It is important to note that the 3D-CNN-AE architectures employed in this
study were not optimized or specifically tuned for the
dataset. Substantial performance gains are likely achievable through
careful architectural design and hyperparameter tuning.

\clearpage

\clearpage
\section{Challenge 2: Forecasting}

\subsection{Introduction}

The fundamental goal of forecasting in fluid mechanics is to predict the future evolution of a flow field from a finite sequence of past observations. Forecasting plays a central role in applications where anticipating the flow state is critical, such as flow control, adaptive sensing, and reduced-order modeling.

Forecasting fluid flows poses distinct challenges arising from the complex, multiscale, and often chaotic nature of fluid motion. Even small modeling or measurement errors can grow rapidly in time, limiting the useful prediction horizon. Nevertheless, many flows exhibit coherent spatio-temporal structures or quasi-periodic components that can be exploited for prediction. For example, cavity flows exhibit Rossiter modes associated with feedback loops between the shear layer and the cavity interior, while pitching airfoils display periodic vortex shedding modulated by the prescribed kinematics. These temporal correlations motivate methods that can identify and model dominant dynamics in an efficient, data-driven manner.


We pose two challenges that each highlight different capabilities and difficulties for forecasting fluid flows. The first dataset we consider is a turbulent cavity flow (\S\ref{sec:forecasting_cavity}), characterized by broadband turbulence with dominant resonant tones, providing a stringent test of short-horizon predictability. The second is a laminar pitching airfoil (\S\ref{sec:forecasting_airfoil}), where the flow evolution is strongly coupled to prescribed kinematics, requiring parametric models that incorporate known input signals such as the instantaneous angle of attack. Together, these problems form a representative pair of forecasting benchmarks for reduced-complexity modeling of fluid flows.

\subsection{Baseline methods}

\subsubsection{Dynamic mode decomposition (DMD)}\label{sec:dmd}

Dynamic mode decomposition (DMD) approximates a linear operator $\bfsym{K}$ that advances the state from one snapshot to the next,
\begin{equation}\label{eqn:koopman}
\tilde{\bfsym{q}}_{k+1} = \bfsym{K}\bfsym{q}_{k},
\end{equation}
where $\bfsym{q}_{k}=\bfsym{q}(k\Delta t)$ denotes the snapshot at time $t=k\Delta t$, so that Eq.~\ref{eqn:koopman} represents propagation over a single time step. Various algorithms for computing DMD exist. Here, we adopt the exact DMD algorithm, known for its robustness and ability to handle non-sequential data where snapshots are provided in pairs. The process begins with two data matrices,
\begin{equation}\label{eqn:datamat}
    \bfsym{Q}_1^{M-1} = \qty[\bfsym{q}_{1},\bfsym{q}_{2},\dots,\bfsym{q}_{M-1}], \quad
   \bfsym{Q}_2^{M} = \qty[\bfsym{q}_{2},\bfsym{q}_{3},\dots,\bfsym{q}_{M}],
\end{equation}
which can either consist of sequential but shifted snapshots or individual snapshot pairs (each pair separated by the same time step, $\Delta t$). The next step is to compute the singular value decomposition (SVD)
\begin{equation}\label{eqn:dmd,svd}
    \bfsym{Q}_1^{M-1} = \bfsym{U}\bfsym{\Sigma}\bfsym{V}^*.
\end{equation}
Retaining only the $N$ largest singular values (and corresponding columns of $\bfsym{U}$ and $\bfsym{V}$) yields a rank-$N$ truncation, $\bfsym{U}_N$, $\bfsym{\Sigma}_N$, $\bfsym{V}_N$,
whose components are then used, together with the shifted data matrix, to form the approximation of the propagation matrix,
\begin{equation}
    \tilde{\bfsym{K}} = \bfsym{U}_N^*\bfsym{Q}_2^{M}\bfsym{V}_N\bfsym{\Sigma}_N^{-1}.
\end{equation}
In the following, we do not explicitly denote the rank truncation and omit the subscript $N$. The eigendecomposition of this matrix,
\begin{equation} \label{eqn:dmdevp}
\tilde{\bfsym{K}}=\bfsym{D}\bfsym{\Lambda}\bfsym{D}^{-1},
\end{equation}
is then used to compute the DMD modes and eigenvalues. Given $\bfsym{D}$, we proceed by computing the matrix of DMD modes,
\begin{equation}\label{eqn:dmd,modes}
    \bfsym{\Phi} = \bfsym{Q}_2^{M}\bfsym{V}\bfsym{\Sigma}^{-1}\bfsym{D}.
\end{equation}
The $i$-th column of $\bfsym{\Phi}$ corresponds to the $i$-th DMD mode, $\bm{\phi}_i$, while the $i$-th diagonal entry of $\bm{\Lambda}$ in Eq. (\ref{eqn:dmdevp}), $\lambda_i$, provides the mode's complex frequency, $\omega_i=\log(\lambda_i)/\Delta t$. Since these frequencies are complex, the temporal dynamics are generally characterized by an oscillatory component and an amplification or damping rate. This distinction makes DMD modes inherently dynamic, unlike POD modes, where temporal dynamics are represented through time-dependent expansion coefficients to describe a specific trajectory. Once $\tilde{\bfsym{K}}\approx\bfsym{K}$ is determined, one could, in principle, use Eq. (\ref{eqn:koopman}) to advance the flow field step by step in time, starting from an initial condition $\bfsym{q}_0$. By leveraging the linearity of Eq. (\ref{eqn:koopman}), the initial condition can be directly propagated to any desired time $t$ by expressing it in the basis of DMD modes and evolving the solution as
\begin{equation}\label{eqn:dmd,evolve}
\tilde{\bfsym{q}}(k\Delta t) = \sum_{i=1}^{N} \bm{\phi}_i \mathrm{e}^{i\bfsym{\omega}_i k\Delta t} b_i,
\end{equation}
where the solution is represented as the sum of the $N$ retained DMD modes, each evolving according to its complex frequency $\omega_i$, starting from the initial amplitude and phase determined by $b_i$. The vector of complex initial amplitudes is obtained by solving the least-squares regression problem
\begin{equation}\label{eqn:dmd,IC}
\bfsym{b} = \bfsym{\Phi}^+\bfsym{q}_0,
\end{equation}
here most conveniently expressed using the pseudoinverse of the DMD mode matrix, $\bfsym{\Phi}^+$, which can be efficiently obtained via SVD.

\subsubsection{Long short-term memory (LSTM) networks}\label{sec:LSTM}

Long short-term memory (LSTM) networks \cite{ref_lstm} are a class of recurrent neural networks (RNNs) specifically designed to learn long-range temporal dependencies in sequential data. They are particularly well-suited for time-series forecasting problems, where the prediction at a given time benefits from information carried forward from many previous time steps. Unlike standard feedforward architectures such as multi-layer perceptrons (MLPs, which are memoryless) and vanilla recurrent neural networks (RNNs, which struggle with vanishing gradients), LSTMs use a gated memory cell to retain and selectively update information over long time horizons, improving learning of long-term temporal dependencies in dynamical forecasts.  In addition to the flow state $\bfsym{q_k}$, an LSTM maintains a hidden state $\bfsym{h}_k$ used for prediction and a cell state $\bfsym{c}_k$ that acts as a longer memory.  These internal states are updated as 
\begin{equation}
(\bfsym{h}_{k+1}, \bfsym{c}_{k+1}) = f_{\theta}(\bfsym{q}_k, \bfsym{h}_k, \bfsym{c}_k)
\end{equation}
and the forecasted flow state is given as
\begin{equation}
\tilde{\bfsym{q}}_{k+1} = g_{\theta}(\bfsym{h}_{k+1}),
\end{equation}
where $f_{\theta}$ and $g_{\theta}$ are the LSTM state-update map and read-out map, respectively.  The network parameters $\theta$ are optimized during training to minimize a predefined loss function, in our case, the mean-squared error of the forecast.  LSTMs have become a standard choice for forecasting complex dynamical systems \citep{vlachas2018data, lindemann2021survey}, including fluid flows \citep{hasegawa2020machine, rahman2019nonintrusive}.

\subsection{Challenge 2.1: Forecasting of a turbulent cavity flow}\label{sec:forecasting_cavity}
\vspace{-0.25cm}
\begin{figure}[H]
 \centering
 \includegraphics[width=1\textwidth,trim={0 2.5in 0 0},clip]{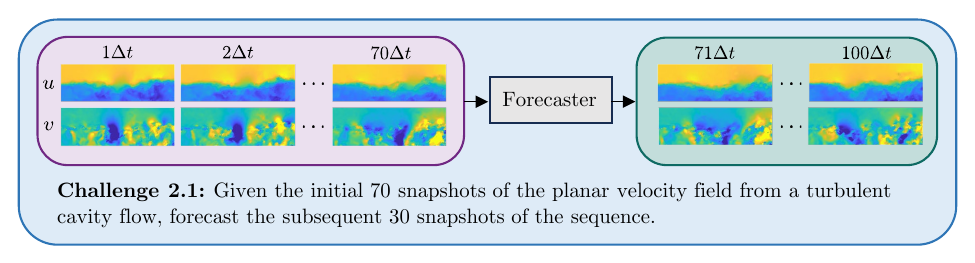}
\end{figure}
\vspace{-0.25cm}

\subsubsection{Data}

This challenge uses experimental data from a high-Reynolds-number flow over an open cavity, obtained using planar time-resolved particle image velocimetry (TR-PIV) by \citet{zhang2020spectral}. This case represents real-world data from a complex, mixed broadband-tonal turbulent flow in the presence of measurement noise. The center plane of the open cavity has a length-to-depth ratio of $L/D=6$ and a width-to-depth ratio of $W/D=3.85$. The Reynolds number based on the cavity depth is $\Re=\rho U_{\infty}D/\mu \approx 3.3\times 10^5$, and the Mach number is $M=U_{\infty}/c_{\infty}=0.6$, where  $c_{\infty}$ is the far-field speed of sound. The sampling rate is $f_s=16$ kHz, and the field of view is resolved by $154\times51$ points. The database comprises 16,000 snapshots of the instantaneous streamwise ($u$) and wall-normal ($v$) velocity. For further details on the experimental setup, refer to \citet{zhang2020spectral}.

\begin{figure}[h!]
 \centering
\includegraphics[width=1\textwidth,trim={0 0cm 0 0},clip]{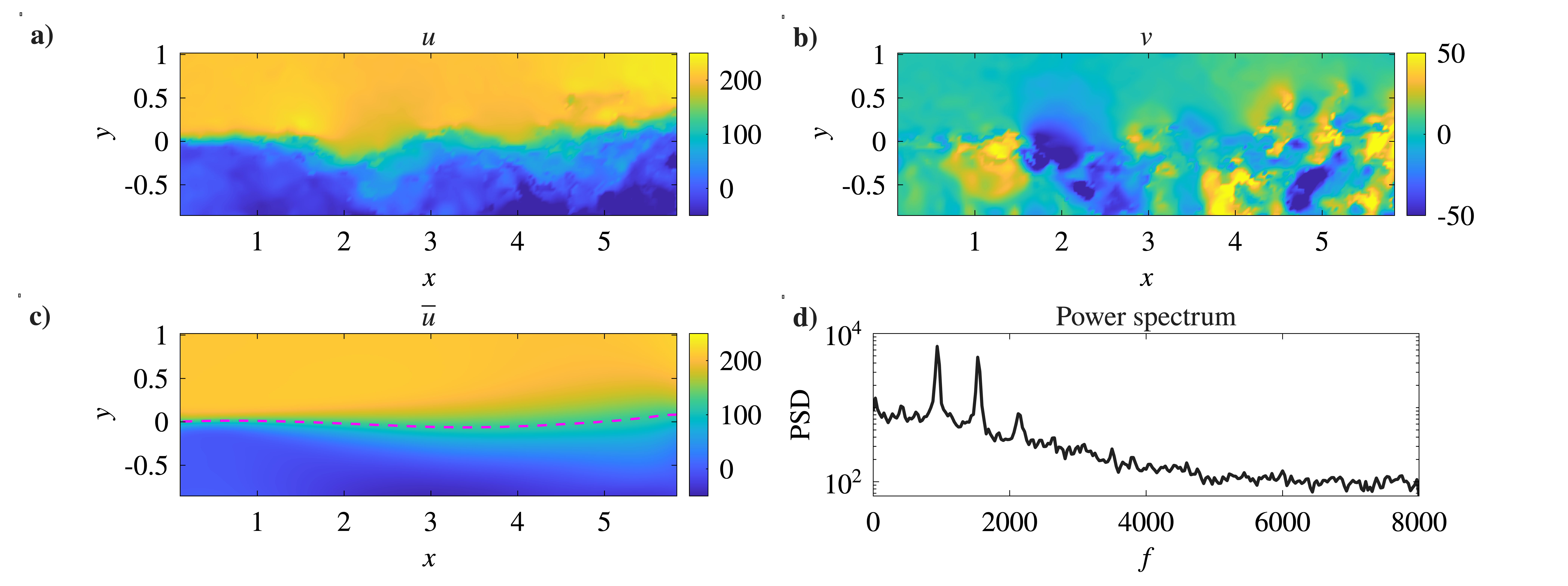}
 \caption{Challenge 2.1: Instantatous velocity field of experimental open cavity flow data by \citet{zhang2020spectral}: (a) streamwise velocity, $u$; (b) wall-normal velocity, $v$; (c) mean streamwise velocity and 120 m/s isoline (dashed magenta line); (d) power spectrum at $(x,y)=(5,0)$. Velocities are in m/s.}
\label{fig:cavityPIV_overview}
\end{figure}

The instantaneous streamwise and wall-normal velocity fields from a representative snapshot are shown in Figure \ref{fig:cavityPIV_overview}(a,b). The flow is evidently turbulent, as confirmed by the broadband-tonal power spectrum in Figure \ref{fig:cavityPIV_overview}(b), computed at $(x,y) = (5,0)$. Superimposed on the turbulent background are distinct tonal peaks corresponding to Rossiter modes, which arise from a resonance between shear-layer instability waves and upstream-traveling acoustic waves. These acoustic waves originate at the trailing edge and synchronize with the shear-layer instabilities, thereby completing the resonance loop \cite{rossiter1964wind}.

The training set contains 12,800 snapshots, and the test set includes the remaining 3,200 snapshots.  This test block is partitioned into 32 sequences of 100 consecutive snapshots. Participants are provided the first 70 snapshots of every test sequence, and the final 30 snapshots are withheld and must be forecast. 

To motivate the challenge, we first estimate the time interval over which realistic predictions can be expected by considering the characteristic time and length scales of the setup. Given a Mach number of 0.6, a cavity length of 0.159 m, a measurement time step of $\Delta t = \tfrac{1}{f_s} = 6.25 \times 10^{-5}$ s, and assuming a speed of sound of 343 m/s, we estimate that a fluid particle traveling at the freestream velocity would pass through the field of view in approximately 12 time steps. However, the dominant energetic structures in this flow are associated with the Rossiter cycle and are linked to the shear-layer dynamics. As a representative velocity scale for these dynamics, the 120 m/s isoline, shown as the magenta dashed line in Figure \ref{fig:cavityPIV_overview}(c), closely follows the center of the shear layer. Assuming this velocity approximates the convection speed of the relevant structures in the shear layer, we estimate a flow-through time of about 21$\Delta t$. This provides a rough upper bound for realistic predictability, suggesting that forecasts beyond $\sim$20 time steps are likely to degrade significantly. That being said, predictive methods may benefit from the inclusion of low-frequency components present in cavity flows, known as centrifugal modes. To account for their potential influence, we allow for a longer prediction horizon.

\subsubsection{Baseline results}

As our classical baseline method, we use DMD, as described in Section \ref{sec:dmd}, to forecast the cavity flow data. We use $M=70$ consecutive snapshots to construct two data matrices according to Eq.~\ref{eqn:datamat}, for example, $\bfsym{Q}_1^{69}$ and $\bfsym{Q}_2^{70}$, and build a DMD model that uses the first 70 snapshots to predict the next 30. It is important to note that the DMD model does not ``learn'' from large datasets; instead, it performs a one-shot forecast based on the data used to compute it. For each of the 32 segments of the testing data, we construct a DMD model following Eqs.~(\ref{eqn:dmd,svd})–(\ref{eqn:dmd,modes}). The initial mode amplitudes $\bfsym{b}$ are then computed via Eq.~\ref{eqn:dmd,IC} and used in Eq.~\ref{eqn:dmd,evolve} to evolve the system forward in time by 30 steps. We retain $N=25$ out of the $M-1=69$ available modes to prevent overfitting.

As our baseline machine‑learning method, we use LSTM regression neural networks as introduced in Section \ref{sec:LSTM}. We use MATLAB’s standard LSTM implementation \cite{MATLAB_LSTM2024}, as it is widely accessible, well-documented, and reproducible. Because direct prediction in physical space is computationally prohibitive, we predict in the space of POD coefficients. Specifically, we perform standard space-only POD for dimensionality reduction and use the POD coefficients as latent variables in the machine‑learning model. The space‑only POD modes are computed from the data matrix containing the full training dataset, that is, $\bfsym{Q}_1^{12,800}$ in the notation of Eq. (\ref{eqn:datamat}). For brevity, we refer to the full training data matrix as $\bfsym{Q}$ and approximate it via the economy SVD,
$
\bfsym{Q}\approx\bfsym{\Phi}_N\bfsym{\Sigma}_N\bfsym{V}_N^T,
$
where $N=25$ and $\bfsym{\Phi}_N$, $\bfsym{\Sigma}_N$, and $\bfsym{V}_N$ contain the $N$ largest singular values and their corresponding left and right singular vectors. The left singular vectors, $\bfsym{\Phi}$, are the POD modes, whereas the right singular vectors, $\bfsym{V}$, provide the temporal POD expansion coefficients $ \bfsym{A}=\bfsym{\Sigma}_N\bfsym{V}_N^T$
that allow us to expand the data in terms of the POD modes as $\bfsym{Q}\approx\bfsym{\Phi}_N\bfsym{A}$. Written as the expansion of individual snapshots $\bfsym{q}_k = \bfsym{q}(k\Delta t)$ at the $k$‑th time instant as the product of spatial POD modes and time‑varying coefficients, $\label{eqn:POD,truncation}
\tilde{\bfsym{q}}_k = \sum_{i=1}^N a_i{(k\Delta t)} \bm{\phi}_i,$
we can forecast the flow field by predicting future expansion coefficients, $\bfsym{a}{(t+\Delta t)}$, $\bfsym{a}{(t+2\Delta t)},\dots$, and reconstructing the state with the precomputed modes. In particular, given the $M=70$ consecutive snapshots of a realization of the flow from the test data, we form the data matrix $\bfsym{Q}_1^{70}$ and compute its expansion coefficients by projecting the data onto the modal basis as
$
\bfsym{A}_1^{70} = \bfsym{\Phi}_N^H\bfsym{Q}_1^{70}.
$
That is, we expand the new test data sequence into the truncated mode basis obtained from the training data. To make predictions from a segment of $M=70$ consecutive snapshots, we form the coefficient matrix $\bfsym{A}_1^{70}$, use $\bfsym{A}_1^{69}$ as the input sequence for the LSTM, and use the shifted $\bfsym{A}_2^{70}$ as the target sequence. From the total of 12,800 snapshots, this yields a training sample of 182 realizations of shifted coefficient matrix pairs to train the LSTM network. Note that the network is trained to predict only the next time instant, $\bfsym{a}{(71\Delta t)}$, from the 70 instances provided (of which only 69 are used, as explained below) and is then applied to the shifted sequence that includes its own prediction to produce a total of 30 forecasted snapshots, that is, the remaining expansion coefficients $\bfsym{a}{(72\Delta t)},\dots,\bfsym{a}{(100\Delta t)}$ required for the challenge. This procedure is often referred to as closed‑loop forecasting.

The network is trained using the standard Adam optimizer with a constant learning rate of 0.001 for 200 epochs, with data shuffled each epoch. We set the mini‑batch size to 26, so the training data are divided into 7 mini‑batches per epoch, resulting in a total of 1400 iterations. The state vector includes both velocity components. The LSTM configuration includes a sequence input layer, one LSTM layer with 128 hidden units, and a fully connected output layer. Following best practices, the coefficients are normalized to zero mean and unit variance, and the mean‑square error is used as the loss function \cite{MATLAB_LSTM2024}. The LSTM was trained on data sequences of 70 snapshots, but because the network must be trained from the shifted sequences $\bfsym{A}_1^{69}$ and $\bfsym{A}_2^{70}$, the effective input sequences have length 69. In principle, the LSTM can learn dependencies spanning the entire 69‑step history, but its effective memory length is determined by what it learns through the gating mechanism. At inference, we initialize the network state with snapshots 2–70 of the testing sequence so that the final input to the network is the true snapshot at time step 70, and the first forecasted step is predicted from the last instance of the testing sequence. This reduces the initial error and subsequent error accumulation. We note this departure from the conventional practice (where, consistent with training, the last entry of the input sequence is predicted) because the cavity data are stationary, allowing the 69‑long input sequence to be shifted arbitrarily.

To make quantitative comparisons between the full flow fields predicted by DMD and LSTM, we use the normalized mean-squared error (NMSE),
\begin{equation}
\label{eq:e_nmse}
e_k = \frac{E\left\{ \|  \tilde{\boldsymbol{q}}_k - \boldsymbol{q}_k \|^2 \right\}}{E\left\{ \|  \boldsymbol{q}_k \|^2 \right\}},
\end{equation}
where the expectation is an ensemble average.

\begin{figure}[h!]
 \centering
\includegraphics[width=1\textwidth,trim={0 0 0 0},clip]{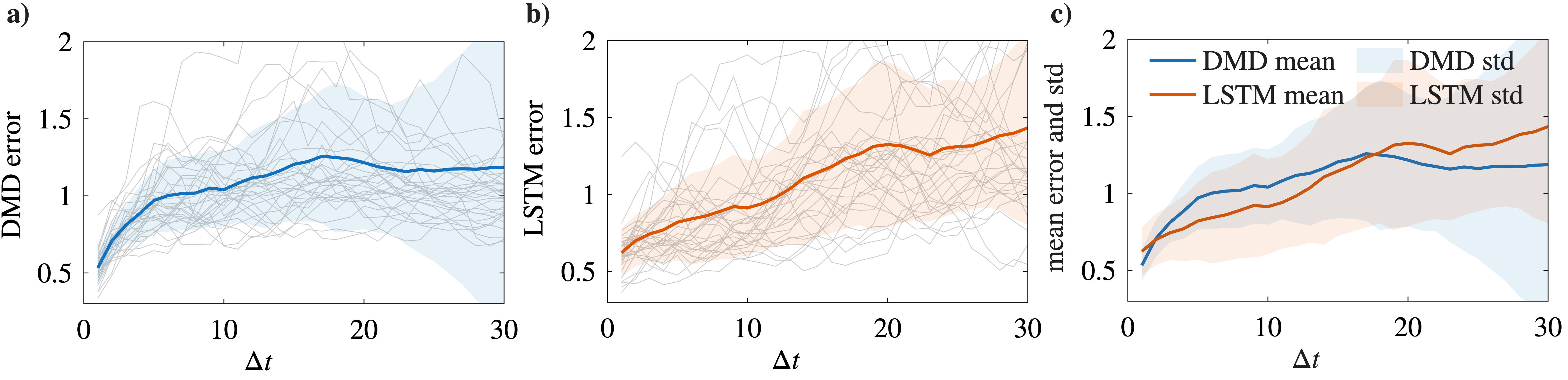}
 \caption{Challenge 2.1: Prediction error of the forecasts. Gray lines show the 32 individual test sequences; the thick line is their mean and the shaded band indicates standard deviation. (a) DMD; (b) LSTM; (c) direct comparison of methods (mean NMSE and standard deviations).}
\label{fig:cavityPIV_err_DMD_vs_LSTM}
\end{figure}
The prediction errors for DMD and LSTM are shown in Figure~\ref{fig:cavityPIV_err_DMD_vs_LSTM}. Individual test-sequence errors (gray curves) display substantial spread for both methods, as reflected by the shaded standard-deviation bands. The mean error increases approximately monotonically at short lead times for both methods and then levels off. The direct comparison in panel~\ref{fig:cavityPIV_err_DMD_vs_LSTM}(c) shows that DMD is slightly more accurate at the first predicted time step, but its error increases more rapidly thereafter, and LSTM attains a lower mean error over the subsequent 16 time steps. However, even the LSTM forecast exceeds an NMSE of one by lead time 13, indicating that the predictive capability of both methods is effectively exhausted beyond this horizon.

\begin{figure}[h!]
 \centering
\includegraphics[width=1\textwidth,trim={0 0 0 0},clip]{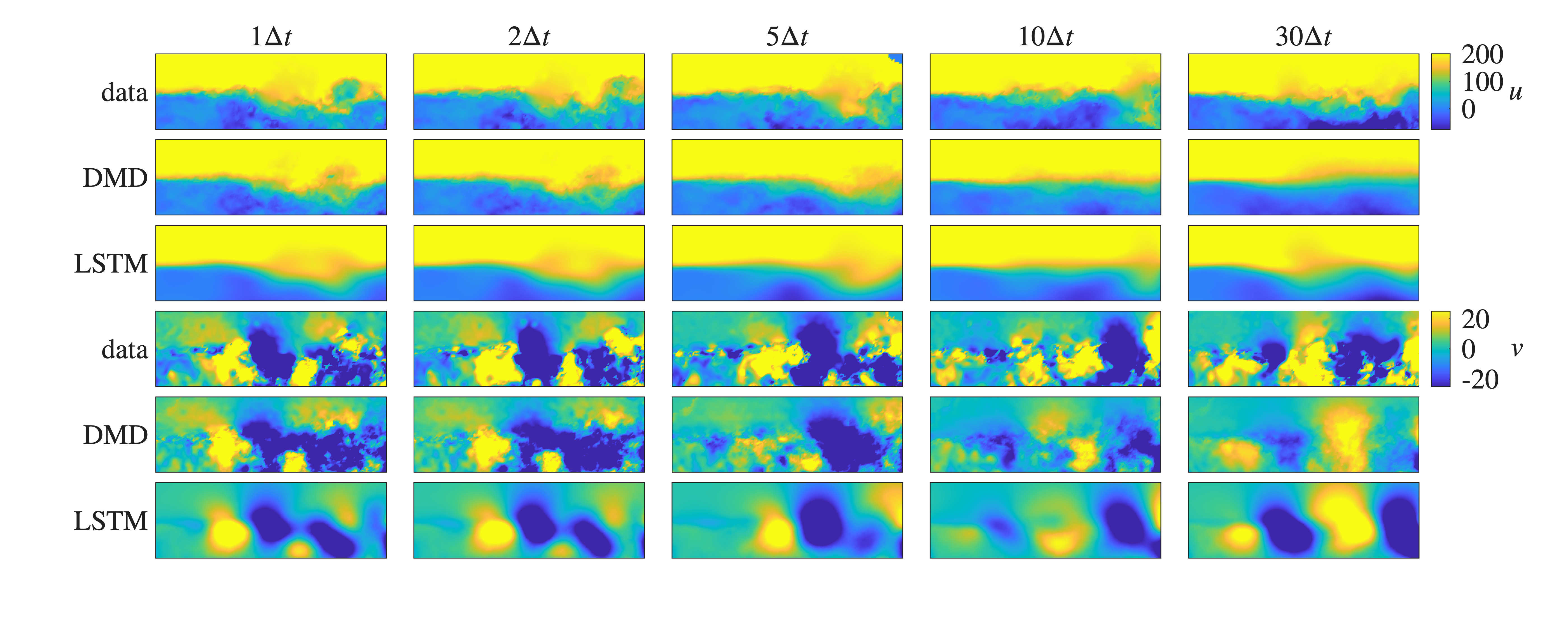}
 \caption{Challenge 2.1: Comparison of instantaneous flow fields for ground truth, DMD, and LSTM predictions. The streamwise velocity $u$ is shown on the top rows and the wall-normal velocity $v$ on the bottom rows. Forecast horizons correspond to the frame immediately after the input sequence and to $+2$, $+5$, $+10$, and $+30$ time steps.}
\label{fig:cavityPIV_prediction_DMD_vs_LSTM}
\end{figure}
Figure \ref{fig:cavityPIV_prediction_DMD_vs_LSTM} shows the $u$- and $v$-velocity components of the reference data (ground truth) and the corresponding DMD and LSTM forecasts at five representative lead times. The most apparent difference between DMD and LSTM is that the DMD forecasts retain much finer turbulent structure, closely resembling the reference data, whereas the LSTM forecasts appear significantly smoother. This is consistent with how the two models are constructed: DMD predictions lie in the linear subspace spanned by the DMD modes, which are themselves built as linear combinations of the snapshot data used to identify the model. By contrast, the LSTM operates in a reduced latent space obtained from a truncated POD representation. Recall that the LSTM is not applied to the raw flow fields but to the POD expansion coefficients; the flow is then reconstructed from a limited number of 25 POD modes, which are optimally averaged spatial structures and therefore inherently smoother than the original data. In other words, we use POD for dimensionality reduction and treat the POD coefficients as latent variables that are advanced in time by the LSTM. In summary, although the DMD forecasts display finer turbulent structure and might therefore be perceived as the more realistic representation of the flow, Figure~\ref{fig:cavityPIV_err_DMD_vs_LSTM} shows that the much smoother LSTM forecasts achieve comparable, and often lower, prediction errors. As a final side note, simply increasing the number of retained POD modes in the LSTM model does not improve accuracy or make the forecasts resemble the DMD predictions or the ground truth more closely. Instead, the higher-dimensional latent space makes training more difficult and convergence less reliable.

\subsection{Challenge 2.2: Forecasting of pitching airfoils with parametric variations}\label{sec:forecasting_airfoil}
\vspace{-0.25cm}
\begin{figure}[H]
 \centering
 \includegraphics[width=1\textwidth,trim={0 0.5in 0 0},clip]{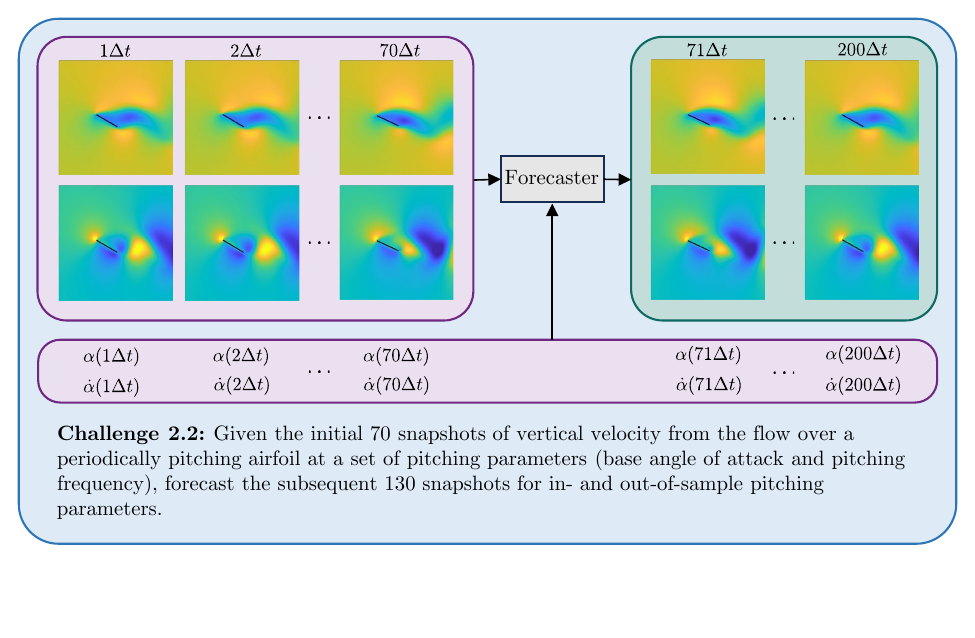}
\end{figure}
\vspace{-0.25cm}

\subsubsection{Data}

This challenge uses DNS data of the flow over a flat plate airfoil undergoing periodic pitching motions about its midchord, with the time-varying angle of attack
\begin{equation}
    \label{eq:periodicpitching}
\alpha(t) = \alpha_0 -  \alpha_P \sin(2\pi f_P t).
\end{equation}
The Reynolds number (based on the freestream velocity and chord length) is 100. The simulations utilize an immersed boundary method \cite{taira:07ibfs, taira:fastIBPM}, and the simulation parameters and numerical setup are identical to those described for the same configuration in Ref.~\cite{towne2023database}. Compared to the tubulent cavity flow in the first forecasting challenge, the dynamics in each simulation are relatively simple due to the low, laminar Reynolds number. The distinguishing features of this dataset that make forecasting nontrivial are the fact that it includes continuous actuation (pitching) at a range of different frequencies, about base states that can be both stable (for $\alpha_0 = 25^\circ$) and unstable with natural vortex shedding (for $\alpha_0 > 27^\circ$). The natural vortex shedding frequency is approximately $f_N = 0.24$ for $\alpha_0 = 30^\circ$, which falls within the range of pitching frequencies considered.

\begin{figure}[h!]
 \centering
\includegraphics[width=0.9\textwidth,trim={0 0cm 0 0},clip]{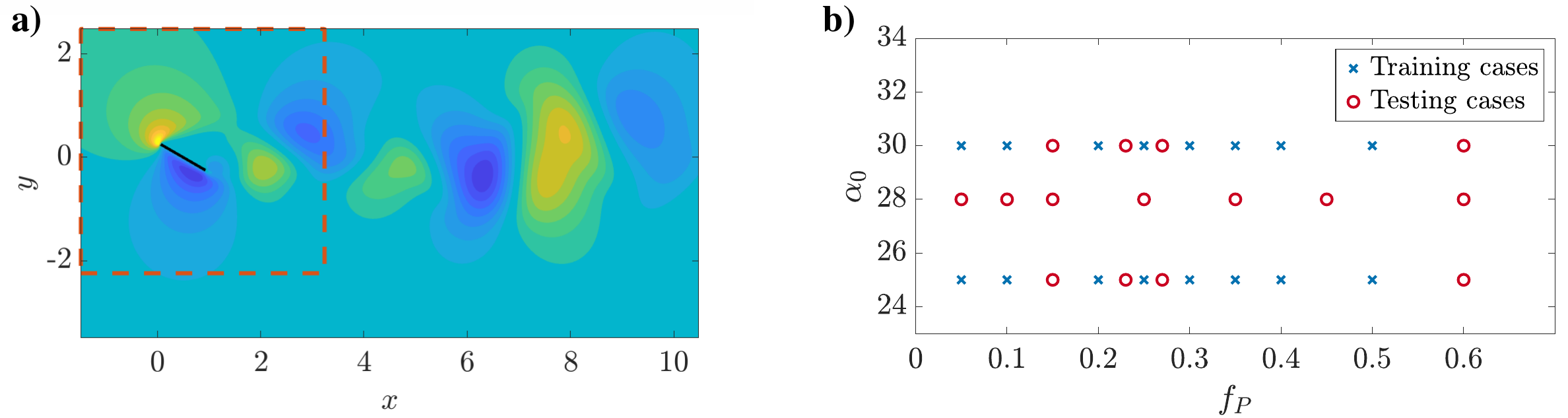}
 \caption{Challenge 2.2: (a) Example flowfield snapshot (vertical velocity, $v$) for the pitching airfoil dataset. The red dashed box indicates the subregion of the domain used for the benchmark problem. (b) Base angles of attack $\alpha_0$ and pitching frequencies used for the training (in sample) and testing (out of sample) datasets.}
\label{fig:airfoilwindow}
\end{figure}

The dataset consists of the instantaneous streamwise ($u$) and transverse ($v$) velocity, collected in a window surrounding the airfoil, as indicated in Fig.~\ref{fig:airfoilwindow}(a). The data are also downsampled relative to the simulations by a factor of four in the $x$ and $y$ directions, giving each snapshot a size of $60\times 60$. Data are collected only after initial transients have decayed. For the purposes of developing models for predicting the evolution of this flowfield, we also assume knowledge of the pitching kinematics, characterized by $\alpha(t)$ and $\dot\alpha(t)$. 

The training dataset consists of 16 different pitching profiles, with two base angles of attack $\alpha_0\in \{25^\circ,30^\circ\}$ and eight pitching frequencies $f_P\in \{0.05,0.1,0.2,0.25,0.3,0.4,0.5 \}$, where this frequency is nondimensionalized by the freestream velocity ($U_\infty$) and airfoil chord ($c$). The pitching amplitude is fixed at $\alpha_P = 5^\circ$. The dataset used for testing consists of these parameters, as well as 15 additional parameter combinations not featured in the training data, as indicated in Fig.~\ref{fig:airfoilwindow}(b). These are referred to as in-sample and out-of-sample cases, respectively. These additional out-of-sample parameter combinations include cases where either one or both of $\alpha_0$ and $f_P$ differ from the training cases, and include a frequency that lies outside of the range used for training.  Each case consists of 800 snapshots collected with timestep $\Delta t= 0.1$ (non-dimensionalized by $c$ and $U_\infty$).

Similar to the turbulent cavity flow example considered in Sec.~\ref{sec:forecasting_cavity}, we aim to identify models capable of forecasting future velocity fields, given an initial sequence of data. This initial sequence (consisting of 70 time steps of data) includes both velocity field and airfoil kinematics ($\alpha(t), \dot\alpha(t)$), with the airfoil kinematics also provided for the forecasting timesteps, as indicated in the Challenge 2.2 schematic. The forecast window consists of 130 timesteps of data. 


\subsubsection{Baseline results}
\label{sec:airfoilresults}


We first train a model for the dynamics using DMD with control, a variant of DMD that enables the incorporation of known input signals into the assumed discrete-time linear dynamics. That is, Eq.~\eqref{eqn:koopman} is modified such that the dynamics are
\begin{equation}\label{eqn:dmdc}
\tilde{\bfsym{q}}_{k+1} = \bfsym{K}\bfsym{q}_{k} + \bfsym{B}\bfsym{u}_{k}.
\end{equation}
This approach was first described in the context of DMD in Ref.~\cite{Proctor:SIAMJADS16}. Here, the input signal $\bfsym{u}_{k}$ consists of the instantaneous angle of attack $\alpha(k\Delta t)$ and pitch rate $\dot\alpha(k \Delta t)$. 
To reduce the dimensionality of the model, we project the data onto the leading $r=20$ POD modes, computed from the concatenation of training data at all pitching frequencies. We emphasize that while the DMD method used for Challenge 2.1 described in Sec.~\ref{sec:forecasting_cavity} identifies a separate model for each testing sequence, here we seek a single global model that can be used for different airfoil kinematics. 

As in Sec.~\ref{sec:forecasting_cavity}, we additionally consider a simple LSTM neural network to perform this forecasting task. We similarly divide each training sequence into segments of 70 snapshots, using a 50\% overlap between adjacent segments, and train a model to predict the state one timestep into the future. The dimensionality of the problem is reduced by projecting the data onto the first 20 POD modes, as was also done for the DMD model. 
The learning rate, number of epochs, and number of LSTM bocks in the single hidden layer are the same as those used in Sec.~\ref{sec:forecasting_cavity}. We similarly quantify the error of the DMD and LSTM models using the NMSE, as given by Eq.~\eqref{eq:e_nmse}.



\begin{figure}[h!]
 \centering
\includegraphics[width=1\textwidth,trim={0 0 0 0},clip]{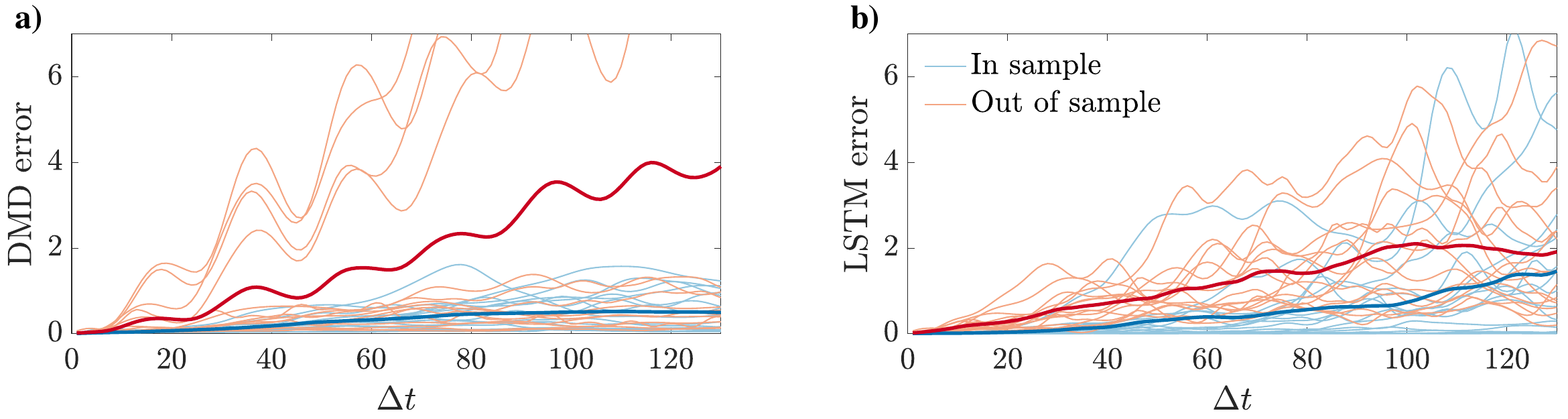}
 \caption{Challenge 2.2: Prediction error (NMSE) of the forecasts for Challenge 2.2, using (a) DMD and (b) LSTM models. Thin lines show individual testing cases, divided between those parameter combinations included in (blue) and excluded from (red) the training dataset, with thick lines showing the average error across each of these two groups.}
\label{fig:airfoil_err_DMD_vs_LSTM}
\end{figure}
Figure~\ref{fig:airfoil_err_DMD_vs_LSTM} shows the error for both the DMD and LSTM models across the forecasting window. The error is partitioned into those cases with parameters used for model training (in sample), and those with different parameters (out of sample). Overall, the LSTM model has a lower total average error for the out-of-sample cases, but a slightly larger average error for the in-sample cases. The large average error for the out-of-sample cases in the DMD model is largely due to the four cases for which the DMD model errors grow rapidly, which correspond to the cases where $\alpha_0 \in \{25^\circ,30^\circ\}$ and $f_P \in \{0.23,0.27\}$, which are all close to the natural vortex shedding frequency of the system. For the in-sample parameter values also used in the training, the average LSTM model error continues to grow over the forecast window, whereas the DMD model error appears to level off after approximately 80 timesteps. 

\begin{figure}[h!]
 \centering
\includegraphics[width=1\textwidth,trim={0 0 0 0},clip]{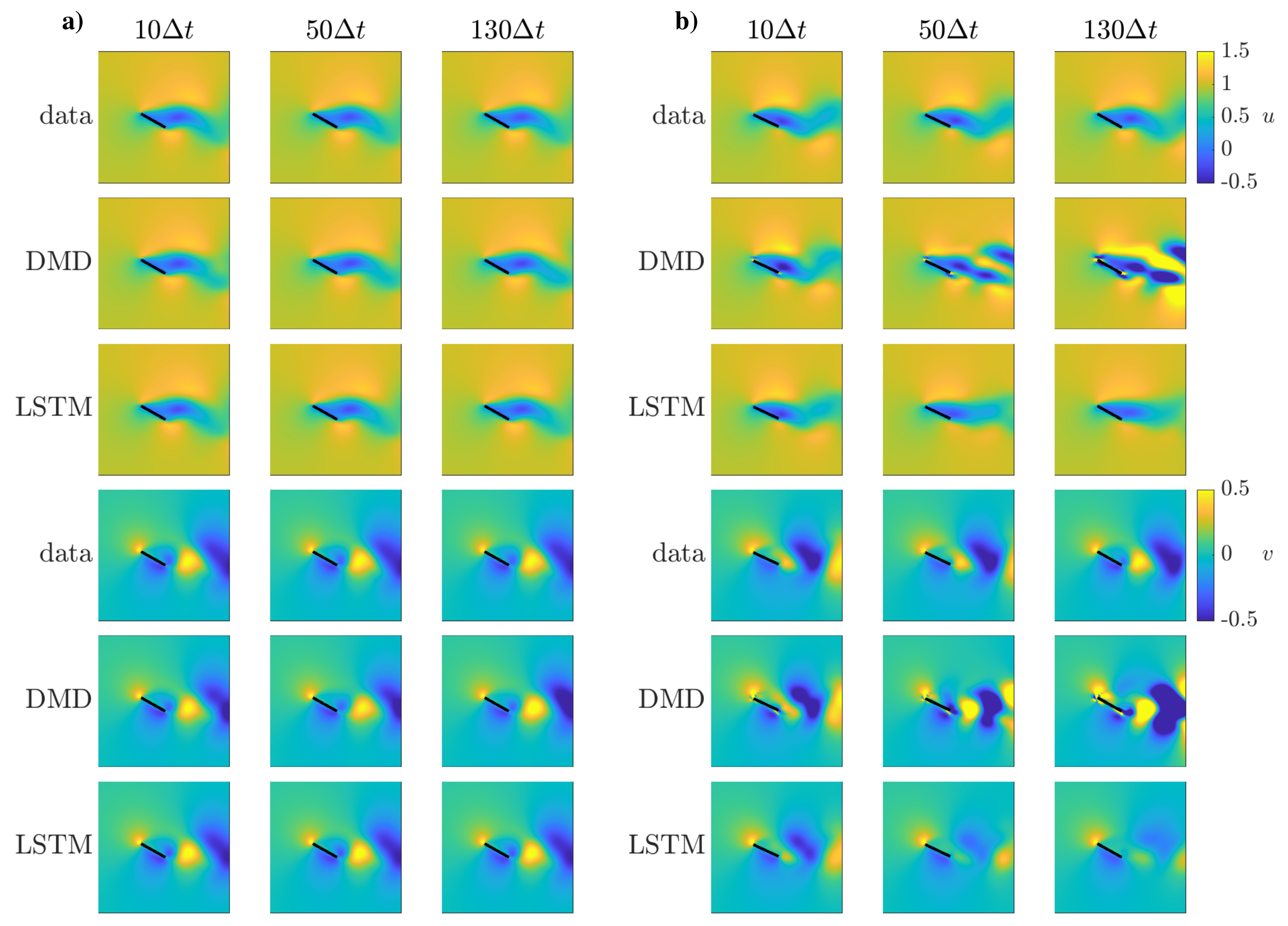}
 \caption{Comparison of instantaneous flow fields for ground truth, DMD, and LSTM predictions for Challenge 2.2, for pitching parameters (a) $\alpha_0 = 30^\circ$, $f_P = 0.25$ and (b) $\alpha_0 = 30^\circ$, $f_P = 0.27$. The streamwise velocity $u$ is shown on the top rows and the wall-normal velocity $v$ on the bottom rows. Columns denote forecast predictions at 10, 50, and 130 timesteps.}
\label{fig:airfoil_prediction_DMD_vs_LSTM}
\end{figure}
Figure~\ref{fig:airfoil_prediction_DMD_vs_LSTM} compares the flowfield predictions of both the DMD and LSTM models for two airfoil pitching parameter combinations. Figure~\ref{fig:airfoil_prediction_DMD_vs_LSTM}(a) shows the case where $\{\alpha_0,f_P\} = \{30^\circ,0.25\}$, which is included in the training set. This pitching frequency is also close to the natural vortex shedding frequency, and the vortex shedding dynamics are observed to lock on to the pitching. Accordingly, the response is periodic and is accurately captured by both models across the entire forecasting window. Figure~\ref{fig:airfoil_prediction_DMD_vs_LSTM}(b) shows the case where $\{\alpha_0,f_P\} = \{30^\circ,0.27\}$, which also exhibits periodic dynamics with vortex shedding phase-locked to pitching. These airfoil kinematics are not included in the training data, and the accuracy of both models is significantly worse. The DMD model predicts velocity fluctuations that grow too large over the forecasting horizon, though it appears to capture the correct phase of the flow field fluctuations (most clear in the $v$ component) across the forecasting window. Note that this was one of the four cases with growing DMD model error in Fig.~\ref{fig:airfoil_err_DMD_vs_LSTM}(a). In contrast, the LSTM model predicts velocity field fluctuations that appear to decay over the forecasting window. While also not capturing the correct periodic dynamics, this results in lower error.

\clearpage


\section{Challenge 3: Sensing}

\subsection{Introduction}

The fundamental goal of sensing in fluid mechanics is to reconstruct or estimate the state of a flow from partial, often sparse, measurement data. Sensing is critical for applications where comprehensive direct measurements are impractical or impossible. For example, in an experiment, particle image velocimetry might provide velocity data but not thermodynamic quantities, with limited resolution and a restricted field of view.  In practical applications, measurements may be available for only a few probes, such as a few pressure sensors on the surface of an aerodynamic body. Fluid flows, particularly turbulence, pose distinct challenges for sensing.  Turbulent flows are chaotic, multi-scale, high-dimensional, and nonlinear. These characteristics limit the information about unseen flow states that can be deduced from limited measurements and violate the assumptions underlying many canonical state-estimation methods.  Broadly speaking, flow sensing is a key capability for the effective monitoring, control, optimization, and understanding of fluid systems.

Using a turbulent jet and a turbulent boundary layer as representative examples, this challenge seeks new data-driven methods that address these challenges and enable improved flow sensing.  In particular, the challenge is designed to probe two limiting cases for the available sensor data.  The jet challenge uses measurements that are spatially sparse but temporally resolved, allowing the time history to be exploited to estimate the current targets. The boundary-layer challenge uses measurements that are not time-resolved but are spatially dense on the wall, allowing spatial patterns to be exploited to estimate the instantaneous velocity field away from the wall.  Due to these differing limits, we use two sets of baseline methods: a Wiener filter and multi-layer perceptron for the jet, and linear stochastic estimation and a convolutional neural network for the boundary layer.

\subsection{Baseline methods}

\subsubsection{Linear stochastic estimation (LSE)} \label{sec:epod}

Linear stochastic estimation (LSE) offers the optimal linear regressor for reconstructing a target signal from correlated but potentially noisy measurements~\cite{adrian1994stochastic}. Given a vector of measurements $\bfsym{y}$ (e.g., wall quantities such as shear stress or pressure) and corresponding target signals $\bfsym{z}$ (e.g., the velocity field), LSE seeks a linear transfer function $\bfsym{T}$ such that the instantaneous estimated target
\begin{equation}
\tilde{\bfsym{z}} = \bfsym{T} \, \bfsym{y}
\end{equation}
minimizes the expected mean-squared error, $E\left\{ \| \tilde{\bfsym{z}}  - \bfsym{z}\|^2 \right\}$.  The optimal transfer function is
\begin{equation}
\bfsym{T} = \bfsym{C}_{zy} \left(\bfsym{C}_{yy} + \bfsym{N} \right)^{-1},
\end{equation}
where $\bfsym{C}_{zy}$ is the cross-correlation between the targets and the measurements, $\bfsym{C}_{yy}$ is the autocorrelation of the measurements, and $\bfsym{N}$ is the autocorrelation of any additive noise within the measurements.  
In practice, both measurements and targets are often projected onto a low-dimensional basis, such as proper orthogonal decomposition (POD), before applying LSE. Within this framework, the temporal coefficients of the measured quantities are linearly mapped into those of the target field using a transfer function constructed from their empirical correlations. This approach is mathematically equivalent to extended proper-orthogonal decomposition (EPOD)~\cite{boree2003epod,encinar2019loglayer}. LSE and EPOD have been applied successfully in turbulent jets~\cite{tinney2008epodjet}, channels~\cite{guastoni_et_al_2021}, as well as other wall-bounded flows~\cite{guastoni2025fcn}, and serve here as a linear benchmark against which more advanced nonlinear estimators, such as convolutional neural networks, can be evaluated.

\subsubsection{Wiener filter}\label{sec:wiener}

The Wiener filter \citep{Wiener1949extrapolation} constitutes the minimum mean-squared-error linear estimator between wide-sense stationary measurements $\bfsym{y}$ and targets $\bfsym{z}$. That is, we seek a linear transfer function $T(\tau)$ such that the estimated target 
\begin{equation}
\label{eq:wiener_conv}
\tilde{\bfsym{z}}(t) = \int \limits_{-\infty}^{\infty} \bfsym{T}(t-\tau) \bfsym{y}(\tau) d \tau
\end{equation}
minimizes the expected mean-squared error.  Whereas LSE (in its typical form) uses only current, instantaneous measurements to estimate the current target, the Wiener filter uses the full measurement history.  The transfer function that minimizes the error is given in the frequency domain as
\begin{equation}
\label{eq:TF_noncausal}
\hat{\bfsym{T}}(\omega) = \bfsym{S}_{zy} \left( \bfsym{S}_{yy} + \bfsym{N} \right)^{-1},
\end{equation}
where $\bfsym{S}_{zy}$ is the cross-spectral density (CSD) between the targets and the measurements, $\bfsym{S}_{yy}$ is the CSD of the measurements, and $\bfsym{N}$ is the CSD of additive noise within the measurements.  The time-domain transfer function $\bfsym{T}(\tau)$ is recovered via an inverse Fourier transform.  

The formulation described so far is non-causal in the sense that future measurements are required to estimate the current target, as expressed by the infinite upper bound of the integral in Eq.~(\ref{eq:wiener_conv}).  This is not an issue for post-processing tasks where all data are simultaneously accessible.  In other sensing tasks, such as system monitoring or closed-loop control, future data is not available.  Attempting to bypass this requirement by simply truncating the non-causal ($\tau<0$) parts of the transfer function forfeits the optimality of the estimator.

The optimal causal transfer function can be derived by restricting the integration limits in~(\ref{eq:TF_noncausal}) to $(-\infty, 0]$ and again minimizing the expected mean-squared error.  The transfer function that minimizes the error is
\begin{equation}
\label{eq:TF_causal}
\hat{\bfsym{T}} = [\bfsym{S}_{zy} ( \bfsym{S}_{yy} + \bfsym{\hat{N}})^{-1}_{-}]_{+} (\bfsym{S}_{yy} + \bfsym{\hat{N}})_{+}^{-1}.
\end{equation}
Here, $[\cdot]_{\pm}$ and $(\cdot)_{\pm}$ correspond to causal and noncausal components as determined by additive and multiplicative Wiener-Hopf decompositions, respectively \citep{hopf1937zusammenhange, noble1958wienerhopf, daniele2007fredholm, Martini2022resolvent}.  Wiener filters are closely related to a number of other methods, including resolvent-based estimators and Kalman filters \cite{Towne2020resolvent, Martini2022resolvent}, and have been applied for flow estimation tasks in a variety of cases inlcuding channels \citep{arun2023toward}, airfoil wakes \citep{jung2025resolvent}, and jets \citep{maia_real-time_2021}.

\subsubsection{Multi-layer perceptron (MLP)}\label{sec:mlp}

The multi-layer perceptron (MLP)~\cite{ref_mlp} is the most basic type of artificial neural network. These networks consist of two or more layers of nodes, with each node connected to every node in both the preceding and the following layers.  Conceptually, the network represents a static map between an input and output space.  Within the context of sensing, the MLP provides a nonlinear map $\boldsymbol{\mathcal{M}}_{\theta}$ between the measurements $\bfsym{y}(t)$ and the targets $\bfsym{z}(t)$, i.e.,
\begin{equation}
\tilde{\bfsym{y}}(t) = \boldsymbol{\mathcal{M}}_{\theta}\left(\bfsym{z}(t)\right).
\end{equation}
The network parameters $\theta$ are again optimized to minimize a desired loss function, in our case, the mean-squared estimation error.  

To use the MLP as an estimator that accounts for measurement history, analogous to the Wiener filter, we redefine the input as a time-delay variable that consists of measurements over a past time interval.  We note that time histories can be more elegantly accounted for using RNNs or LSTMs, among other advanced architectures; we use an MLP here to provide the simplest possible nonlinear extension of the Wiener filter, which should, in theory, reproduce the Wiener filter if the network is constructed with linear activation functions.  Several examples of application of MLPs to predict fluid flows are presented in Ref.~\cite{srinivasan_et_al}.

\subsubsection{Convolutional neural network (CNN)} \label{sec:cnn_new}

Convolutional neural networks (CNNs)~\cite{lecun} are widely used in applications where the input and output domains share structural similarities, allowing the exploitation of spatial correlations in the data~\cite{ref_segmentation}. In the context of flow sensing, structured input might take the form of two-dimensional measurements (e.g., distributions of wall pressure or wall-shear stress), and structured outputs might include flow fields at target wall-normal locations (e.g., planar velocity fields). Similar to the MLP, the CNN provides a nonlinear map $\boldsymbol{\mathcal{C}}_{\theta}$ between the measurements $\bfsym{y}(t)$ and the targets $\bfsym{z}(t)$, i.e.,
\begin{equation}
\tilde{\bfsym{y}}(t) = \boldsymbol{\mathcal{C}}_{\theta}\left(\bfsym{z}(t)\right).
\end{equation}
The distinguishing feature of an MLP is that the map takes the form
\begin{equation}
\boldsymbol{\mathcal{C}}_\theta = \rho_L \circ \kappa_L \circ \cdots \circ \rho_1 \circ \kappa_1,
\end{equation}
where each $\kappa_\ell$ is a convolutional operator, and each $\rho_\ell$ is a pointwise nonlinearity.  That is, a CNN estimates the target $\bfsym{z}$ from measurement data $\bfsym{y}$ by repeatedly applying local, translation-equivariant convolutions with shared weights. This structure drastically reduces the number of network parameters $\theta$ and leverages the spatial relationship between nearby measurements, making CNNs effective when $\bfsym{y}$ has spatial structure.  A detailed discussion of CNNs and their applications can be found in Ref.~\cite{ref_ccn26}, and a comparison with LSE for flow sensing can be found in Ref.~\cite{nakamura2022lseNN}.

\subsection{Challenge 3.1: Sensing of a turbulent jet using sparse measurements}
\vspace{-0.25cm}
\begin{figure}[H]
 \centering
 \includegraphics[width=1\textwidth,trim={0 0.75in 0 0},clip]{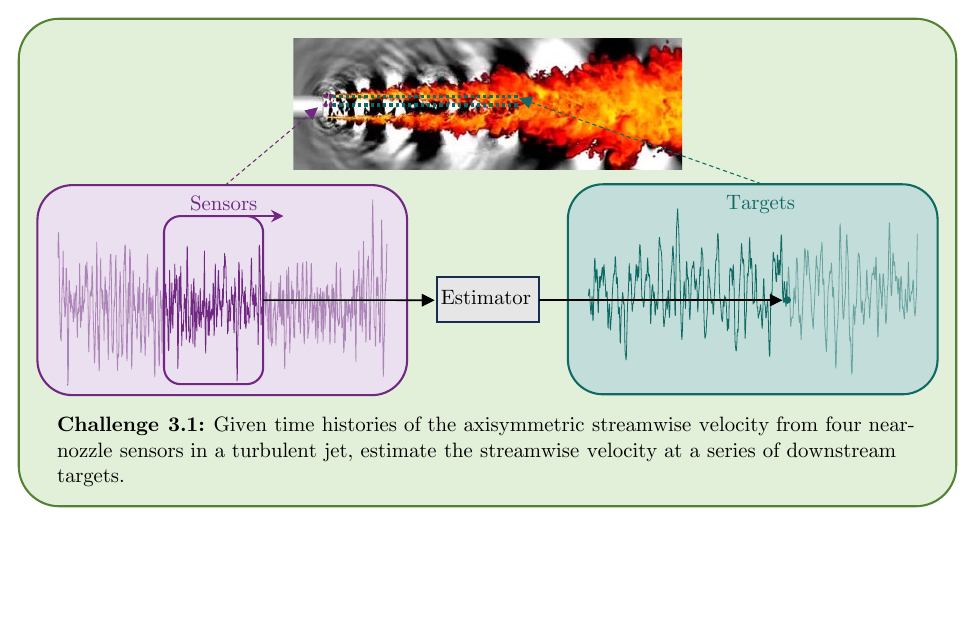}
\end{figure}
\vspace{-0.25cm}

\subsubsection{Data}

This challenge uses large-eddy-simulation (LES) data for a turbulent jet issuing from a contoured convergent-straight nozzle \cite{BresJFM2018}.  The jet Mach number is $M_j = U_j/c_j = 0.9$ and the jet is isothermal ($T_j/T_\infty=1.0$), where $U$ is the mean streamwise velocity, $c$ is the speed of sound, $T$ is the mean temperature, and the subscripts $j$ and ${\infty}$ refer to the jet exit and freestream conditions, respectively.  The Reynolds number is $Re_j = \rho_j U_j D/\mu_j \approx 1\times10^6$, where $D$ is the nozzle exit diameter, $\rho$ is density, and $\mu$ is dynamic viscosity.  The simulation was conducted using the compressible flow solver {\it{CharLES}} \cite{BresAIAAJ2017}, which solves the spatially filtered compressible Navier–Stokes equations on unstructured grids using a finite volume method. The grid used here contains approximately 16 million control volumes.  The LES methodologies, numerical setup, and comparisons with measurements are described in Ref. \cite{BresJFM2018}. The full database \cite{towne2023database} contains 10,000 snapshots of the density, velocity, and pressure fields sampled every 0.2 acoustic time units ($t c_{\infty}/D$). 

The sensor data for this challenge correspond to measurements of the axisymmetric streamwise velocity taken from the LES data at four probe locations, $(x/D,r/D) = (0.25,0), (0.5,0), (0.25,0.5), (0.5,0.5)$.  To mimic the finite size and bandwidth of real sensors, each measurement is defined as a Gaussian-weighted spatial average, which also has the effect of filtering high-frequency components of the true velocity field.  The Gaussians are centered at the four aforementioned positions and have variance $\sigma/D = 0.1$.  Since we are working with the axisymmetric component of the velocity field, each sensor with $r/D > 0$ would be physically implemented as an azimuthal ring of sensors with sufficient density to isolate the azimuthal mean.  These sensor data are to be used to estimate the axisymmetric streamwise velocity at a series of target locations along the jet center line $(r/D=0)$ and lip line $(r/D = 0.5)$.  For consistency with the sensors, each target is also defined using the same Gaussian-weighted average.  The streamwise locations of the targets span $x/D \in [0.5, 10]$ with spacing $\Delta x/D = 0.25$.  

The training data consists of 8,000 snapshots of the sensor and target data, while the test data contains the remaining 2,000 snapshots.  Participants are provided the sensor test data, while the target test data are withheld and must be estimated.  A requirement of this challenge is that methods must be causal in the sense that they require only past and current data to estimate the current state of the targets.  To accommodate methods that use an interval of the previous time history to produce the current estimate, the first 200 snapshots of the sensor test data are excluded when computing the estimation error.

\subsubsection{Baseline results}

To construct the causal Wiener filter using a history of 200 snapshots, cross-spectra were computed using Welch’s method with 400 snapshots per block, 90\% overlap, and a sixth-order sine window \citep{martini2020accuratefrequencydomainidentification}.  While the data are not explicitly corrupted with noise, we assume that each measurement signal contains white noise at an amplitude of 0.1\% to regularize the transfer functions.  

The MLP estimator was likewise constructed to minimize the MSE loss function using a history of 200 snapshots of the sensor readings as input.  To maximize the available training samples, an overlap of 199 was used to define each temporal block, resulting in 9800 samples.  15\% of these training samples were withheld for validation, giving 8330 training and 1470 validation samples in total. Both the input and target data were mean-subtracted and normalized by their standard deviation prior to training.  We considered two different architectures, each with an input layer, three fully connected hidden layers with 512 nodes, and a fully connected output layer.  The first MLP uses linear activation functions, and therefore should theoretically reproduce the Wiener filter.  The second MLP uses a nonlinear (hyperbolic tangent) activation function followed by a 20\% dropout layer to enhance convergence.  For both MLPs, the minibatch size was 64, the learning rate was $0.0001$, and training was performed using an Adam optimizer.  The model was found to be highly prone to overfitting, and we therefore use early stopping, i.e., the iteration with the minimum validation loss.

\begin{figure}[hbt!]
\centering
\includegraphics[width=1\textwidth]{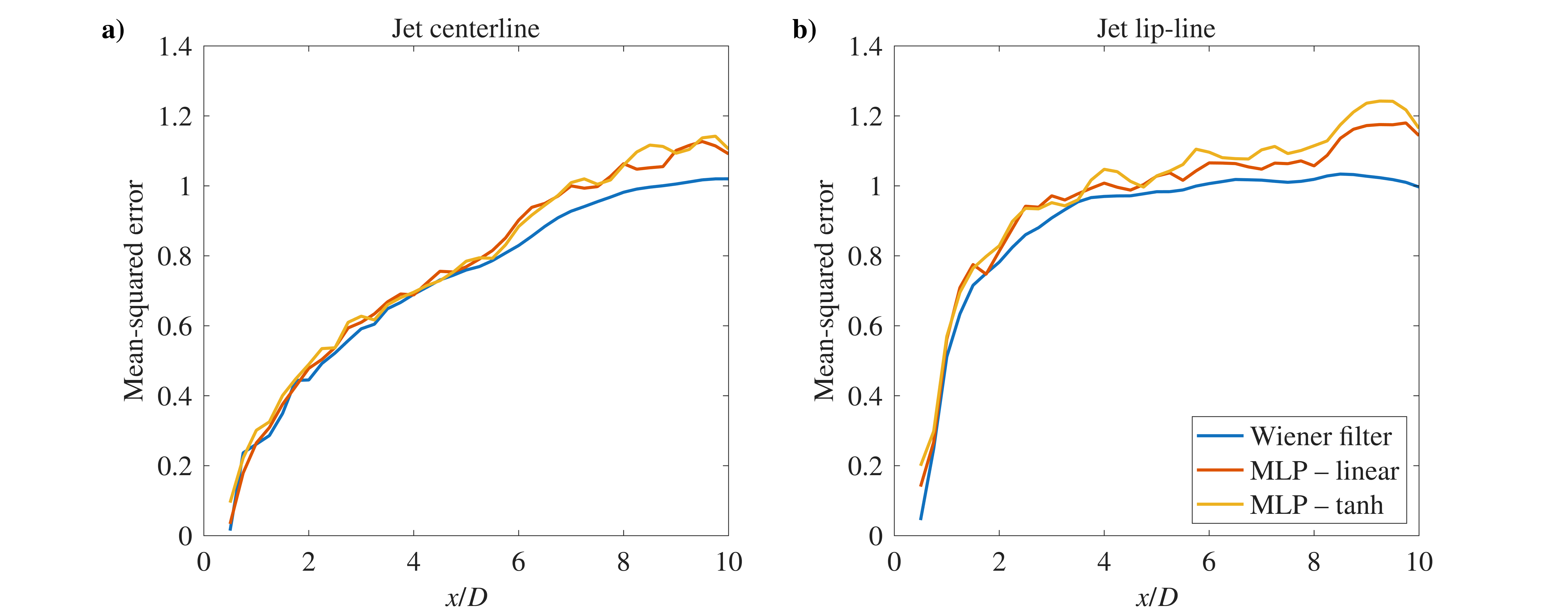}
\caption{Challenge 3.1: Normalized mean-squared error as a function of the streamwise position $x/D$ along the jet (a) centerline and (b) lip line for a Wiener filter, linear MLP, and nonlinear MLP.   \label{fig:wiener_causal}}
\end{figure}

The performance of the estimator is evaluated using the normalized mean-squared error (NMSE),
\begin{equation}
e = \frac{E\left\{ \|  \tilde{\boldsymbol{z}}(t) - \boldsymbol{z}(t) \|^2 \right\}}{E\left\{ \|  \boldsymbol{z}(t) \|^2 \right\}},
\label{eq:NMSE_sensing}
\end{equation}
where the expectation is a time average.  Results for the causal Wiener filter and the two MLPs are shown in Figure~\ref{fig:wiener_causal} as a function of $x/D$ along the jet centerline (a) and lip-line (b).  A striking, and perhaps unexpected, feature of these results is that the nonlinear MLP does not outperform the Wiener filter and, in fact, produces higher errors for most targets.  This highlights the need for sensing algorithms that better leverage patterns within the available time-series data.  Similarly, the linear MLP does not faithfully recover the Wiener filter, i.e., its approximation of the theoretical Wiener filter is less accurate than that obtained using CSDs computed by Welch’s method, as described above.  

In all cases, the error increases with $x/D$ as the distance between the sensors and targets increases.  The error is higher along the lip-line than the centerline; the former is turbulent for all streamwise positions, while the latter is contained within the comparatively laminar potential core up until around $x/D=5.5$.  Indeed, there is a clear slope discontinuity in the centerline error at this location for all three estimators, after which the centerline error quickly reaches similar order-one levels as the lip-line error.

\clearpage

\subsection{Challenge 3.2: Sensing of a turbulent boundary layer from wall stress}
\vspace{-0.25cm}
\begin{figure}[h!]
 \centering
 \includegraphics[width=1\textwidth,trim={0 1.75in 0 0},clip]{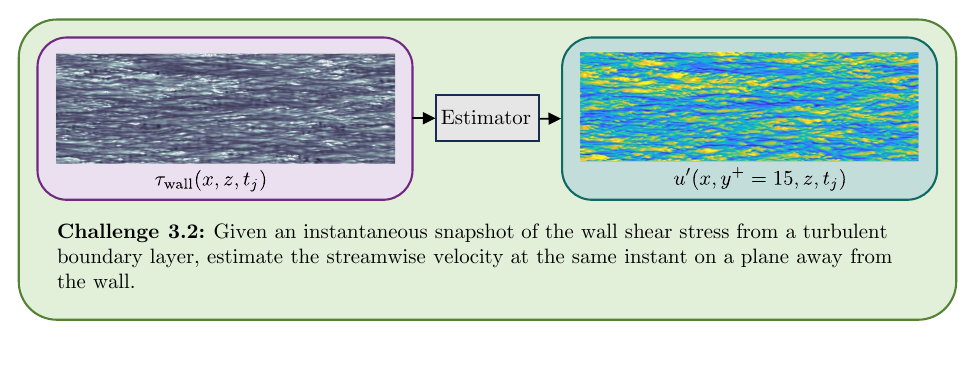}
\end{figure}
\vspace{-0.25cm}

\subsubsection{Data}

This challenge uses data from a direct numerical simulation (DNS) of a zero-pressure-gradient incompressible turbulent boundary layer, specifically the BL1 case described in Ref.~\citep{towne2023database}.  The full database contains 10,000 three-dimensional flowfields sampled every $\Delta t^{+} \approx 1.5$ and spanning a physical time of 26 eddy turnovers.  

From this database, the sensor data for this challenge corresponds to the streamwise wall-shear stress fluctuation $\tau_{w}$ in a streamwise-spanwise ($x$–$z$) plane covering friction Reynolds numbers in the range $Re_{\tau} \approx 000\text{--}000$ and discretized using $507 \times 150$ grid points.  The target data consists of the streamwise velocity fluctuation at a wall-normal distance $y^+=15$ within the same $x$–$z$ domain, where $y^+=y u_{\tau} / \nu$ is the inner-scaled wall-normal coordinate, $u_{\tau}=\sqrt{\overline{\tau}_{w} / \rho}$ is the friction velocity, $\overline{\tau}_{w}$ is the mean streamwise wall-shear stress, and is $\rho$ the fluid density.  This is an important problem in experiments, where real-time wall measurements are less intrusive than introducing a probe to measure the flow in the near-wall region, particularly at high Reynolds numbers.

The training data consists of 1,000 snapshots of the sensor and target data, while the test data contains an additional 250 snapshots.  Participants are provided with the sensor test data, while the target test data are withheld and must be estimated. A requirement of this challenge is that a particular snapshot of the target velocity field is to be estimated using only the snapshot of the wall-shear stress sensor reading at the same instant in time.

\subsubsection{Baseline results}

The LSE estimator is constructed by using the training data to approximate the cross-correlation between the targets and measurements and the autocorrelation of the measurements.  Although the DNS data are not explicitly corrupted with noise, we assume that the sensor data contain white noise at an amplitude of 0.1\%
to regularize the transfer functions.

The CNN architecture comprises 17 convolutional layers each with a stride of 1 and the same padding, so the spatial resolution is preserved throughout while only the channel dimension is transformed. The number of filters increases over the first 8 layers from the initial value of 4 up to 32, then symmetrically decreases over the next 8 layers back to 1, forming an hourglass-shaped hierarchy of feature maps that enables multi-scale representation and subsequent reconstruction of the target field. All intermediate layers use \texttt{tanh} activation to provide nonlinear transformations appropriate for signed flow quantities, whereas the final layer uses a \texttt{linear} activation to produce unconstrained regression outputs. Training is performed with an adaptive learning rate schedule that reduces the learning rate when the validation loss plateaus, combined with early stopping and checkpointing, so that the final deployed model is the one that attains the minimum validation loss, thereby limiting overfitting while retaining good generalization performance.

\begin{figure}[hbt!]
\centering
\includegraphics[width=\textwidth]{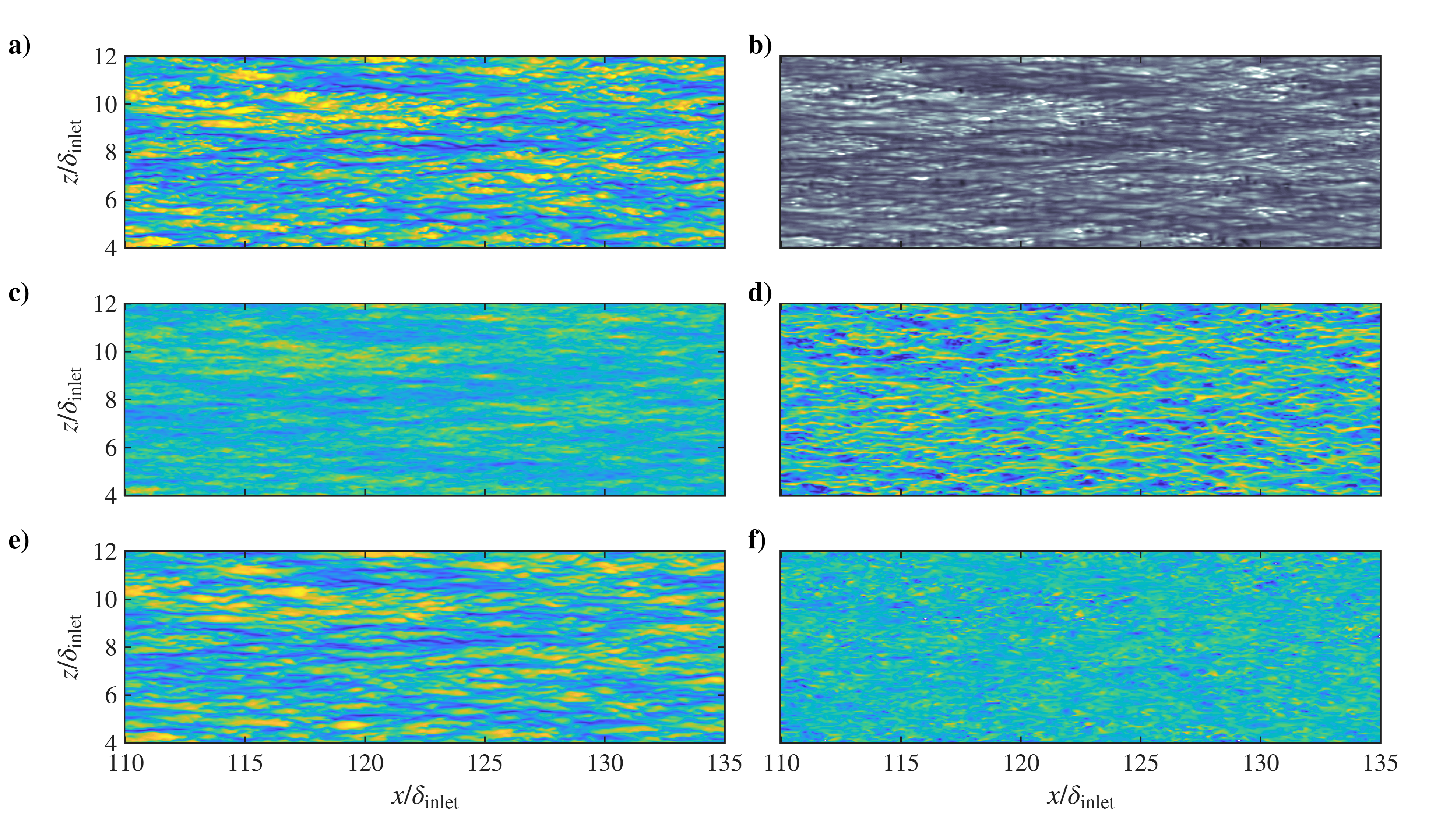}
\caption{Challenge 3.2: A snapshot of the target, sensor, and error fields. (a) the streamwise velocity fluctuation at $y^{+} = 15$ from DNS; (b) the wall shear stress from DNS; (c) LSE estimate of the velocity field; (d) difference between LSE and DNS velocity fields; (e) CNN estimate of the velocity field; (f) difference between CNN and DNS velocity fields. \label{fig:BL_sensing}}
\end{figure}

Figure~\ref{fig:BL_sensing}(a) and (b) show one snapshot of the streamwise velocity fluctuation at $y^{+} = 15$, i.e., the target to be sensed, and the wall shear stress, i.e., the sensor reading from the DNS.  The corresponding estimates of the target velocity field produced on the LSE and CNN, and the difference compared to the DNS, are shown in Figure~\ref{fig:BL_sensing}(c-f). While both estimates show the presence of near-wall streaky structures, the LSE-estimated field significantly underpredicts their amplitude.  Additionally, the CNN-estimated field contains more of the small-scale structures present in the DNS field. The NMSE from Eq.~\ref{eq:NMSE_sensing}, the metric both methods were designed to minimize, averaged over the test data, is 0.912 and 0.239 for the LSE and CNN estimators, respectively.  

\section{Participation Guidelines}\label{sec:participation}
\vspace{\baselineskip}

The objective of this challenge—to facilitate direct comparisons between data-driven methods for compression, forecasting, and sensing of turbulent flows—will only be achieved with broad community participation.  The challenge is open to anyone who wishes to participate, and we invite you to do so.  Participation guidelines are summarized below; additional details, as well as periodic updates, can be found on the challenge webpage: \url{fluids-challenge.engin.umich.edu}.

{\bf Data.}  All data required for the challenge are available on the webpage in a series of HDF5 files.  Training data are provided for each challenge.  Additionally, input data required for testing the methods are provided for the forecasting and sensing challenges, i.e., initial conditions and/or parameters and sensor measurements, respectively.  Additional test data are withheld to ensure blind tests, as described in more detail below.

{\bf Code.}  The website also provides example code for the baseline methods for each challenge.  In addition to allowing you to reproduce the baseline results included in this paper, this code serves as a template for methods submitted to the challenge.  Each baseline method has a main driver script that reads the training data (and the input test data if applicable), executes the method (by calling additional functions in a corresponding utilities directory), and writes out a results file.  You are expected to release open-source code for each method submitted to the challenge, following this same structure; code that does not run or is not complient with the required format will not be accepted.  The code can be written in any language; indeed, we used multiple languages (a mix of MATLAB and Python) for the baseline methods to emphasize this point.  We also provide an evaluation script for each challenge that will be used to calculate performance metrics for each submitted method, as described below.

{\bf Blind testing.}  To ensure fair comparisons, we will use blind tests to evaluate each method.  The main driver script (described above) for each method must write out a results file containing the same information as included in the example baseline code, e.g., the estimated target values for the sensing challenges.  You will send these results files to the relevant challenge point of contact (POC, see the webpage for details).  For the forecasting and sensing challenges, the results file strictly contains data, i.e., predictions of the future flow state and targets, respectively.  For the compression challenges, the results file will contain the encoder and decoder themselves.  The POC will then execute the evaluation function, which reads both the results file and the necessary (withheld) test data, and writes a metrics file that will be sent back to you. 

{\bf AIAA Journal Virtual Collection and invited presentations.}  In addition to the open-source code described above, the final deliverable of the challenge will be a Virtual Collection (analogous to a special issue) in the AIAA Journal.  Specifically, for each challenge you complete, you will submit an article describing the proposed method and reporting its results, as contained in the metrics file supplied by the POC, and compared to the provided baseline results.  You are also welcome to optionally include results for additional test problems of your choosing that are relevant to a given challenge type.  If you apply more than one method to the same challenge, they should be included in a single submission, and the methods should be appropriately compared and contrasted, in addition to comparisons with the baseline methods.  If you complete multiple challenges, you should submit multiple articles.  Each submission will undergo the standard AIAA Journal review process and should follow all AIAA Journal guidelines.  Given the expected scope, we anticipate relatively brief articles.  The publishability of a submission is not contingent on the performance of the method(s).  Indeed, we encourage negative results to achieve a balanced and comprehensive overview of available methods and the current state of the art.   Instead, the review process will focus on the quality of the presentation and compliance with the challenge requirements.  Finally, at the conclusion of the challenge, we will write an overview article summarizing the submission to and results of the challenge.  Additionally, we will host a special session at the AIAA SciTech and/or Aviation meetings, during which a subset of participants will be invited to present their methods and results.  

{\bf Administration.}  The challenge webpage contains, in addition to the data and example codes described above, additional details not included in this summary, including the POCs for each challenge.  We request that you indicate your intention to participate by completing the survey on the webpage.  In addition to helping us prepare, you will be added to a mailing list to ensure that you receive relevant updates.  Finally, the webpage also outlines the challenge timeline and key dates and contains instructions for submitting your results.  We look forward to receiving your submissions!



\section*{Funding Sources}

OTS was funded in part by AFOSR grant no. FA9550-22-1-0541, in support of Challenge 2.1. AT was funded in part by ONR grant no. N00014-22-1-2561, in support of Challenge 3.1. 
STMD was funded in part by DOE grant DE-SC0025597, in support of Challenge 2.2.
ALD was funded in part by NSF grant no. 2317254 and NASA grant no. 80NSSC23K1498, in support of Challenges 1.1 and 1.2. 

\section*{Acknowledgments}

We thank Dr. Junoh Jung for providing parts of the Wiener–Hopf code used in Challenge 3.1 and Prof. Andr\'es Botella for his contributions to the CNN code used in Challenge 3.2. 

\bibliography{challenge}

@inproceedings{rumsey2022data,
  title={Data-Driven Turbulence Modeling: Summary and Outcomes of the 2022 NASA Symposium},
  author={Rumsey, Chris and Coleman, Gary},
  booktitle={High-Fidelity Large-Eddy Simulations (LES) and Direct Numerical Simulations (HiFiLeD) Symposium 2022},
  year={2022}
}

@article{MOULT2005285,
title = {A decade of CASP: progress, bottlenecks and prognosis in protein structure prediction},
journal = {Current Opinion in Structural Biology},
volume = {15},
number = {3},
pages = {285-289},
year = {2005},
note = {Sequences and topology/Nucleic acids},
issn = {0959-440X},
doi = {10.1016/j.sbi.2005.05.011},
author = {John Moult}
}

@inproceedings{wang2018glue,
  title={GLUE: A multi-task benchmark and analysis platform for natural language understanding},
  author={Wang, Alex and Singh, Amanpreet and Michael, Julian and Hill, Felix and Levy, Omer and Bowman, Samuel},
  booktitle={Proceedings of the 2018 EMNLP workshop BlackboxNLP: Analyzing and interpreting neural networks for NLP},
  pages={353--355},
  year={2018},
  doi = {10.18653/v1/W18-5446}
}

@inproceedings{deng2009imagenet,
  title={{ImageNet}: A large-scale hierarchical image database},
  author={Deng, Jia and Dong, Wei and Socher, Richard and Li, Li-Jia and Li, Kai and Fei-Fei, Li},
  booktitle={2009 IEEE conference on computer vision and pattern recognition},
  pages={248--255},
  year={2009},
  organization={IEEE},
  doi = {10.1109/CVPR.2009.5206848}
}

@article{kutz2025acceleratingscientificdiscoverycommon,
      title={Accelerating scientific discovery with the common task framework}, 
      author={J. Nathan Kutz and Peter Battaglia and Michael Brenner and Kevin Carlberg and Aric Hagberg and Shirley Ho and Stephan Hoyer and Henning Lange and Hod Lipson and Michael W. Mahoney and Frank Noe and Max Welling and Laure Zanna and Francis Zhu and Steven L. Brunton},
      year={2025},
      journal={arXiv:2511.04001},
      doi={10.48550/arXiv.2511.04001} 
}

@article{yagoubi2024neurips,
  title={Neurips 2024 ml4cfd competition: Harnessing machine learning for computational fluid dynamics in airfoil design},
  author={Yagoubi, Mouadh and Danan, David and Leyli-Abadi, Milad and Brunet, Jean-Patrick and Mazari, Jocelyn Ahmed and Bonnet, Florent and Farjallah, Asma and Cinnella, Paola and Gallinari, Patrick and Schoenauer, Marc and others},
  journal={arXiv preprint arXiv:2407.01641},
  year={2024},
  doi = {10.48550/arXiv.2407.01641}
}

@article{ohana2024well,
  title={The well: a large-scale collection of diverse physics simulations for machine learning},
  author={Ohana, Ruben and McCabe, Michael and Meyer, Lucas and Morel, Rudy and Agocs, Fruzsina and Beneitez, Miguel and Berger, Marsha and Burkhart, Blakesly and Dalziel, Stuart and Fielding, Drummond and others},
  journal={Advances in Neural Information Processing Systems},
  volume={37},
  pages={44989--45037},
  year={2024}
}

@misc{MATLAB_LSTM2024,
  author = {MathWorks},
  title = {Time series forecasting using deep learning},
  howpublished = {\url{https://www.mathworks.com/help/deeplearning/ug/time-series-forecasting-using-deep-learning.html}},
  note = {Accessed: 2025-03-13},
  year = {2024}
}

@article{towne2023database,
  title={A database for reduced-complexity modeling of fluid flows},
  author={Towne, Aaron and Dawson, Scott TM and Br{\`e}s, Guillaume A and Lozano-Dur{\'a}n, Adri{\'a}n and Saxton-Fox, Theresa and Parthasarathy, Aadhy and Jones, Anya R and Biler, Hulya and Yeh, Chi-An and Patel, Het D and others},
  journal={AIAA journal},
  volume={61},
  number={7},
  pages={2867--2892},
  year={2023},
  doi={10.2514/1.J062203}
}

@Article{rossiter1964wind,
  author    = {Rossiter, J. E.},
  journal   = {RAE Technical Report No. 64037},
  title     = {Wind tunnel experiments on the flow over rectangular cavities at subsonic and transonic speeds},
  year      = {1964},
  publisher = {Ministry of Aviation; Royal Aircraft Establishment; RAE Farnborough}
}

@Article{zhang2020spectral,
  author    = {Zhang, Y. and Cattafesta, L. N. and Ukeiley, L.},
  journal   = {Experiments in Fluids},
  title     = {Spectral analysis modal methods (SAMMs) using non-time-resolved PIV},
  year      = {2020},
  number    = {11},
  pages     = {1--12},
  volume    = {61},
  doi = {10.1007/s00348-020-03057-8}
}

@article{ref_mlp,
  author    = {D. E. Rumelhart and G. E. Hinton and R. J. Williams},
  title     = {Learning internal representations by error propagation},
  institution = {California University at San Diego, La Jolla Institute for Cognitive Science},
  year      = {1985},
  number    = {ICS-8506},
  address   = {San Diego, CA}
}

@article{ref_lstm,
  author    = {Hochreiter, Sepp and Schmidhuber, Jürgen},
  title     = {Long Short-Term Memory},
  journal   = {Neural Computation},
  volume    = {9},
  number    = {8},
  pages     = {1735--1780},
  year      = {1997},
  publisher = {MIT Press},
  doi = {10.1162/neco.1997.9.8.1735}
}

@article{ref_segmentation,
  author    = {J. Long and E. Shelhamer and T. Darrell},
  title     = {Fully Convolutional Networks for Semantic Segmentation},
  booktitle = {Proceedings of the IEEE Conference on Computer Vision and Pattern Recognition},
  pages     = {3431--3440},
  year      = {2015},
  doi = {10.1109/CVPR.2015.7298965}
}

@article{lecun,
  author    = {Y. LeCun and L. Bottou and Y. Bengio and P. Haffner},
  title     = {Gradient-based Learning Applied to Document Recognition},
  journal   = {Proceedings of the IEEE},
  volume    = {86},
  pages     = {2278--2324},
  year      = {1998},
  doi = {10.1109/5.726791}
}

@book{ref_ccn26,
  author    = {I. Goodfellow and Y. Bengio and A. Courville},
  title     = {Deep Learning},
  publisher = {The MIT Press},
  year      = {2016}
}

@article{guastoni_et_al_2021,
  author    = {L. Guastoni and A. Güemes and A. Ianiro and S. Discetti and P. Schlatter and H. Azizpour and R. Vinuesa},
  title     = {Convolutional-network models to predict wall-bounded turbulence from wall quantities},
  journal   = {Journal of Fluid Mechanics},
  volume    = {928},
  pages     = {A27},
  year      = {2021},
  doi = {10.1017/jfm.2021.812}
}

@article{srinivasan_et_al,
  author    = {P. A. Srinivasan and L. Guastoni and H. Azizpour and P. Schlatter and R. Vinuesa},
  title     = {Predictions of turbulent shear flows using deep neural networks},
  journal   = {Physical Review Fluids},
  volume    = {4},
  pages     = {054603},
  year      = {2019},
  doi = {10.1103/PhysRevFluids.4.054603}
}

@article{boree2003epod,
  author    = {Bor{\'e}e, J.},
  title     = {Extended proper orthogonal decomposition: a tool to analyse correlated events in turbulent flows},
  journal   = {Experiments in Fluids},
  volume    = {35},
  number    = {2},
  pages     = {188--192},
  year      = {2003},
  doi = {10.1007/s00348-003-0656-3}
}

@article{encinar2019loglayer,
  author    = {Encinar, M. P. and Jim{\'e}nez, J.},
  title     = {Logarithmic-layer turbulence: a view from the wall},
  journal   = {Physical Review Fluids},
  volume    = {4},
  number    = {11},
  pages     = {114603},
  year      = {2019},
  doi = {10.1103/PhysRevFluids.4.114603}
}

@article{tinney2008epodjet,
  author    = {Tinney, C. E. and Ukeiley, L. S. and Glauser, M. N.},
  title     = {Low-dimensional characteristics of a transonic jet. Part 2. Estimate and far-field prediction},
  journal   = {Journal of Fluid Mechanics},
  volume    = {615},
  pages     = {53--92},
  year      = {2008},
  doi = {10.1017/S0022112008003601}
}

@article{guastoni2025fcn,
  author    = {Guastoni, Luca and Balasubramanian, Arivazhagan G. and Foroozan, Firoozeh and G{\"u}emes, Alejandro and Ianiro, Andrea and Discetti, Stefano and Schlatter, Philipp and Azizpour, Hossein and Vinuesa, Ricardo},
  title     = {Fully convolutional networks for velocity-field predictions based on the wall heat flux in turbulent boundary layers},
  journal   = {Theoretical and Computational Fluid Dynamics},
  year      = {2025},
  volume    = {39},
  pages     = {13},
  doi = {10.1007/s00162-024-00732-y}
}

@article{BresAIAAJ2017,
	author = {Br\`es, G. A.  and Ham, F. E. and Nichols, J. W.  and Lele, S. K.},
	title = {Unstructured Large Eddy Simulations of Supersonic Jets},
	year = {2017},
	journal = {AIAA Journal},
	volume = {55},
	number = {4},
	pages = {1164--1184},	
	doi = {10.2514/1.J055084}
}

@article{BresJFM2018,
  author	= {Br\`es, G. A. and  Jordan, P. and Jaunet, V. and Le Rallic, M. and Cavalieri, A. V. G. and Towne, A. and Lele, S. K. and Colonius, T.  and Schmidt, O. T. },
  title		= {Importance of the nozzle-exit boundary-layer state in subsonic turbulent jets},
  journal	= {Journal of Fluid Mechanics},
  volume={851},
  pages={83--124},
  year		= {2018},
  doi = {10.1017/jfm.2018.476}}

@article{Towne2020resolvent,
  title={Resolvent-based estimation of space--time flow statistics},
  author={Towne, A. and Lozano-Dur{\'a}n, A. and Yang, X.},
  journal={Journal of Fluid Mechanics},
  volume={883},
  year={2020},
  doi = {10.1017/jfm.2019.854}
  }

@article{Martini2022resolvent, 
title={Resolvent-based tools for optimal estimation and control via the {W}iener–{H}opf formalism}, 
volume={937},
journal={Journal of Fluid Mechanics}, 
author={Martini, Eduardo and Jung, Junoh and Cavalieri, André V.G. and Jordan, Peter and Towne, Aaron}, 
year={2022}, 
pages={A19},
doi = {10.1017/jfm.2022.102}
}

@book{Wiener1949extrapolation,
  title     = {Extrapolation, Interpolation, and Smoothing of Stationary Time Series},
  author    = {Wiener, Norbert},
  year      = {1949},
  publisher = {MIT Press},
  address   = {Cambridge, MA}
}

@article{hopf1937zusammenhange,
  author    = {Hopf, Eberhard},
  title     = {{\"U}ber die Zusammenh{\"a}nge zwischen linearen Operationen und verallgemeinerten harmonischen Funktionen},
  journal   = {Mathematische Annalen},
  volume    = {114},
  number    = {1},
  pages     = {161--186},
  year      = {1937},
  publisher = {Springer},
  doi = {10.1007/BF01594171}
}

@book{noble1958wienerhopf,
  author    = {Noble, Ben},
  title     = {Methods Based on the Wiener-Hopf Technique for the Solution of Partial Differential Equations},
  year      = {1958},
  publisher = {Pergamon Press}
}

@article{daniele2007fredholm,
  author    = {Daniele, V. and Lombardi, G.},
  title     = {Fredholm factorization of Wiener–Hopf scalar and matrix kernels},
  journal   = {Radio Science},
  volume    = {42},
  number    = {06},
  pages     = {RS6S01:1--19},
  year      = {2007},
  doi       = {10.1029/2007RS003673}
}

@misc{martini2020accuratefrequencydomainidentification,
      title={Accurate Frequency Domain Identification of ODEs with Arbitrary Signals}, 
      author={Eduardo Martini and André V. G. Cavalieri and Peter Jordan and Lutz Lesshafft},
      year={2020},
      eprint={1907.04787},
      archivePrefix={arXiv},
      primaryClass={eess.SP},
      doi={10.48550/arXiv.1907.04787}, 
}

@article{adrian1994stochastic,
  author    = {Adrian, Ronald J.},
  title     = {Stochastic estimation of conditional structure: a review},
  journal   = {Applied Scientific Research},
  volume    = {53},
  number    = {3-4},
  pages     = {291--303},
  year      = {1994},
  doi       = {10.1007/BF00849100}
}

@article{nakamura2022lseNN,
  author    = {Nakamura, Taku and Fukami, Kai and Fukagata, Koji},
  title     = {Identifying key differences between linear stochastic estimation and neural networks for fluid flow regressions},
  journal   = {Scientific Reports},
  volume    = {12},
  number    = {1},
  pages     = {3863},
  year      = {2022},
  doi       = {10.1038/s41598-022-07515-7}
}

@article{taira:fastIBPM,
	author = {T. Colonius and K. Taira},
	journal = {Computer Methods in Applied Mechanics and Engineering},
	pages = {2131-2146},
	title = {A fast immersed boundary method using a nullspace approach and multi-domain far-field boundary conditions},
	volume = {197},
	year = {2008},
	doi = {10.1016/j.cma.2007.08.014}
	}

@article{taira:07ibfs,
	author = {K. Taira and T. Colonius},
	journal = {Journal of Computational Physics},
	number = 2,
	pages = {2118-2137},
	title = {The immersed boundary method: a projection approach},
	volume = 225,
	year = 2007,
	doi = {10.1016/j.jcp.2007.03.005}
	}

@article{Proctor:SIAMJADS16,
  author = {Proctor, Joshua L. and Brunton, Steven L. and Kutz, J. Nathan},
  title = {Dynamic Mode Decomposition with Control},
  journal = {SIAM Journal on Applied Dynamical Systems},
  volume = {15},
  number = {1},
  pages = {142-161},
  year = {2016},
  doi = {10.1137/15M1013857}
}

@article{sirovich1987turbulence,
  title={Turbulence and the dynamics of coherent structures. I. Coherent structures},
  author={Sirovich, Lawrence},
  journal={Quarterly of Applied Mathematics},
  volume={45},
  number={3},
  pages={561--571},
  year={1987},
  doi = {10.1090/qam/910462}
}

@article{towne2018spectral,
  title={Spectral proper orthogonal decomposition and its relationship to dynamic mode decomposition and resolvent analysis},
  author={Towne, Aaron and Schmidt, Oliver T and Colonius, Tim},
  journal={Journal of Fluid Mechanics},
  volume={847},
  pages={821--867},
  year={2018},
  doi = {10.1017/jfm.2018.283}
}

@book{lumley1970stochastic,
  title={Stochastic tools in turbulence},
  author={Lumley, J. L.},
  year={1970},
  address   = {New York},
  publisher={Academic Press}}

@article{berkooz1993proper,
  title={The proper orthogonal decomposition in the analysis of turbulent flows},
  author={Berkooz, Gal and Holmes, Philip and Lumley, John L},
  journal={Annual Review of Fluid Mechanics},
  volume={25},
  number={1},
  pages={539--575},
  year={1993},
  doi={10.1146/annurev.fl.25.010193.002543}
}

@article{rowley2017model,
  title={Model reduction for flow analysis and control},
  author={Rowley, Clarence W and Dawson, Scott TM},
  journal={Annual Review of Fluid Mechanics},
  volume={49},
  number={1},
  pages={387--417},
  year={2017},
  doi={10.1146/annurev-fluid-010816-060042}
}

@incollection{colonius2025modal,
	series = {Computation and {Analysis} of {Turbulent} {Flows}},
	title = {Chapter 2 - {Modal} decomposition},
    booktitle = {Data Driven Analysis and Modeling of Turbulent Flows},
	isbn = {978-0-323-95043-5},
	publisher = {Academic Press},
	author = {Colonius, T. and Towne, A.},
	editor = {Duraisamy, Karthik},
	year = {2025},
    number = {},
	doi = {10.1016/B978-0-32-395043-5.00008-5},
	pages = {27--81},
}

@incollection{rumelhart1986learning,
  author    = {Rumelhart, David E. and Hinton, Geoffrey E. and Williams, Ronald J.},
  title     = {Learning representations by back-propagating errors},
  booktitle = {Parallel Distributed Processing: Explorations in the Microstructure of Cognition, Volume 1: Foundations},
  editor    = {Rumelhart, David E. and McClelland, James L. and the PDP Research Group},
  publisher = {MIT Press},
  address   = {Cambridge, MA},
  year      = {1986},
  pages     = {318--362}
}

@article{hinton2006reducing,
  author  = {Hinton, Geoffrey E. and Salakhutdinov, Ruslan R.},
  title   = {Reducing the dimensionality of data with neural networks},
  journal = {Science},
  year    = {2006},
  volume  = {313},
  number  = {5786},
  pages   = {504--507},
  doi     = {10.1126/science.1127647}
}

@inproceedings{masci2011stacked,
  title={Stacked convolutional auto-encoders for hierarchical feature extraction},
  author={Masci, Jonathan and Meier, Ueli and Cire{\c{s}}an, Dan and Schmidhuber, J{\"u}rgen},
  booktitle={21st International Conference on Artificial Neural Networks},
  pages={52--59},
  year={2011},
  doi          = {10.1007/978-3-642-21735-7_7}
}

@article{lee2020model,
title = {Model reduction of dynamical systems on nonlinear manifolds using deep convolutional autoencoders},
journal = {Journal of Computational Physics},
volume = {404},
pages = {108973},
year = {2020},
doi = {10.1016/j.jcp.2019.108973},
author = {Kookjin Lee and Kevin T. Carlberg}
}

@article{solera2024beta,
  title={$\beta$-variational autoencoders and transformers for reduced-order modelling of fluid flows},
  author={Solera-Rico, Alberto and Sanmiguel Vila, Carlos and G{\'o}mez-L{\'o}pez, Miguel and Wang, Yuning and Almashjary, Abdulrahman and Dawson, Scott TM and Vinuesa, Ricardo},
  journal={Nature Communications},
  volume={15},
  number={1},
  pages={1361},
  year={2024},
  doi={10.1038/s41467-024-45578-4}
}

@article{vlachas2018data,
  title={Data-driven forecasting of high-dimensional chaotic systems with long short-term memory networks},
  author={Vlachas, Pantelis R and Byeon, Wonmin and Wan, Zhong Y and Sapsis, Themistoklis P and Koumoutsakos, Petros},
  journal={Proceedings of the Royal Society A: Mathematical, Physical and Engineering Sciences},
  volume={474},
  number={2213},
  pages={20170844},
  year={2018},
  doi={10.1098/rspa.2017.0844}
}

@article{lindemann2021survey,
  title={A survey on long short-term memory networks for time series prediction},
  author={Lindemann, Benjamin and M{\"u}ller, Timo and Vietz, Hannes and Jazdi, Nasser and Weyrich, Michael},
  journal={Procedia Cirp},
  volume={99},
  pages={650--655},
  year={2021},
  doi={10.1016/j.procir.2021.03.088}
}

@article{hasegawa2020machine,
  title={Machine-learning-based reduced-order modeling for unsteady flows around bluff bodies of various shapes},
  author={Hasegawa, Kazuto and Fukami, Kai and Murata, Takaaki and Fukagata, Koji},
  journal={Theoretical and Computational Fluid Dynamics},
  volume={34},
  number={4},
  pages={367--383},
  year={2020},
  doi={10.1007/s00162-020-00528-w}
}

@article{rahman2019nonintrusive,
  title={Nonintrusive reduced order modeling framework for quasigeostrophic turbulence},
  author={Rahman, Sk M and Pawar, Suraj and San, Omer and Rasheed, Adil and Iliescu, Traian},
  journal={Physical Review E},
  volume={100},
  number={5},
  pages={053306},
  year={2019},
  publisher={APS},
  doi={10.1103/PhysRevE.100.053306}
}

@article{arun2023toward,
  title = {Towards real-time reconstruction of velocity fluctuations in turbulent channel flow},
  author = {Arun, Rahul and Bae, H. Jane and McKeon, Beverley J.},
  journal = {Physical Review Fluids},
  volume = {8},
  issue = {6},
  pages = {064612},
  numpages = {42},
  year = {2023},
  month = {Jun},
  publisher = {American Physical Society},
  doi = {10.1103/PhysRevFluids.8.064612}
}

@article{maia_real-time_2021,
	title = {Real-time reactive control of stochastic disturbances in forced turbulent jets},
	volume = {6},
	issn = {2469-990X},
	doi = {10.1103/PhysRevFluids.6.123901},
	language = {en},
	number = {12},
	urldate = {2024-07-12},
	journal = {Physical Review Fluids},
	author = {Maia, I. A. and Jordan, P. and Cavalieri, A. V. G. and Martini, E. and Sasaki, K. and Silvestre, F. J.},
	year = {2021},
	pages = {123901}
}

@article{jung2025resolvent,
  title={Resolvent-based estimation of a turbulent wake},
  author={Jung, Junoh and Towne, Aaron},
  journal={arXiv:2507.18837},
  year={2025},
  doi={10.48550/arXiv.2507.18837}
}

\end{document}